\documentclass[journal,compsoc, 9pt]{IEEEtran}

\usepackage[utf8]{inputenc}
\usepackage[T1]{fontenc}
\usepackage[margin=0.75in]{geometry}
\usepackage{amsmath,amssymb}
\usepackage{graphicx}
\usepackage{xurl}
\usepackage{booktabs}
\usepackage{tabularx}
\usepackage[numbers,sort&compress]{natbib}
\usepackage{xcolor}
\usepackage{hyperref}

\hypersetup{
    colorlinks=false,
    pdfborder={0 0 0}
}

\title{Towards a Trustworthy and Governable Internet of Agentics: An Architectural Perspective on Protocols, Systems, and Services}

\author{Haoyi Xiong, \emph{Senior Member, IEEE}, Yi Liu, Jiang Bian, \emph{Member, IEEE}, Peng Diao, Bin Zhu, Tingting Hu, Yi Wang, and Sumi Helal, \emph{Fellow, IEEE}\thanks{H. Xiong, Y. Liu, J. Bian, P. Diao, B. Zhu, T. Hu, and Y. Wang are with Microsoft Corporation. S. Helal is with the Department of Computer Science and Engineering  at the University of Bologna.}}

\begin{document}

\IEEEtitleabstractindextext{%

\begin{abstract}
Large language model applications are increasingly being organized as tool-using and multi-agent workflows rather than as isolated chat interfaces. This perspective paper examines the architectural problem that follows from that shift: agents may reason, invoke tools, and delegate work, but dependable deployment still requires interoperability across frameworks, runtimes, data boundaries, and organizational control domains. We study this problem through the lens of the \emph{Internet of Agentics} (IoA), an emerging service-computing view in which heterogeneous agents are treated as governable service participants rather than only as stand-alone model endpoints. The paper organizes the landscape as a three-layer progression: from today's fragmented agent ecosystem, to interoperable agents, and ultimately to trustworthy and governable autonomy. First, it examines the protocol stack centered on the Agentic AI Foundation, including MCP for agent--tool interaction; A2A for agent--agent coordination; AGENTS.md and Skills-style procedural manifests; and discovery, transport, and trust mechanisms such as AGNTCY and SLIM. Second, it reviews the systems abstractions that make agents deployable as service components, including logical agent architectures, orchestration patterns, gateways, registries, runtime control loops, and hardware-aware inference infrastructure. Third, it situates these developments within services computing, arguing that agentic systems reinterpret classical concerns from SOA and microservices---discovery, composition, governance, pricing, and trust---at the level of cognitive capabilities rather than deterministic functions. The paper concludes that the central IoA research challenge is trustworthy autonomy: ensuring  that interoperable agent services are observable, auditable, privilege-bounded, economically controllable, and sufficiently dependable for real-world production use.
\end{abstract}

}
\maketitle

\section{Introduction}
\label{sec:intro}

Recent agentic systems have moved beyond single-turn assistants toward networks of specialized agents that plan, delegate, invoke tools, and interact with enterprise services~\cite{wang2024survey,guo2024large,contemporary_agent2025,yao2023react,shinn2023reflexion,hong2024metagpt}. This shift changes the principal systems challenge. Model capability remains necessary, but production deployment increasingly depends on whether heterogeneous agents can interoperate across frameworks, tools, data boundaries, and organizational control domains~\cite{wang2024survey,guo2024large,mast2025failures,google2025scaling,wu2023autogen,chase2024langgraph}. Existing deployments remain constrained by fragmented interfaces, duplicated integration logic, weak governance mechanisms, and brittle coordination loops~\cite{wang2024survey,guo2024large,mast2025failures,google2025scaling,wu2023autogen,chase2024langgraph}. The central claim of this paper is that these constraints define a service-computing problem: interoperability must mature into trustworthy autonomy before agentic systems can operate as dependable production services.

We use the term \emph{Internet of Agentics} (IoA) to denote an emerging distributed service environment in which heterogeneous agents communicate, coordinate, and expose capabilities through open protocols rather than isolated proprietary integrations~\cite{ioa2025fundamentals,aaif2025launch,anthropic2024mcp,a2a2025spec}. From a services-computing perspective, the IoA does not displace established service abstractions; instead, it extends them to systems whose components exhibit planning, tool use, memory, and adaptive delegation. The unit of composition shifts from deterministic software services to cognitive agent services, while enduring concerns such as interface description, discovery, orchestration, state management, trust, and accountability remain central~\cite{asc2025survey,papazoglou2003soa,fielding2000rest,masterman2024landscape}.

\begin{figure}[t]
\centering
\includegraphics[width=0.95\linewidth]{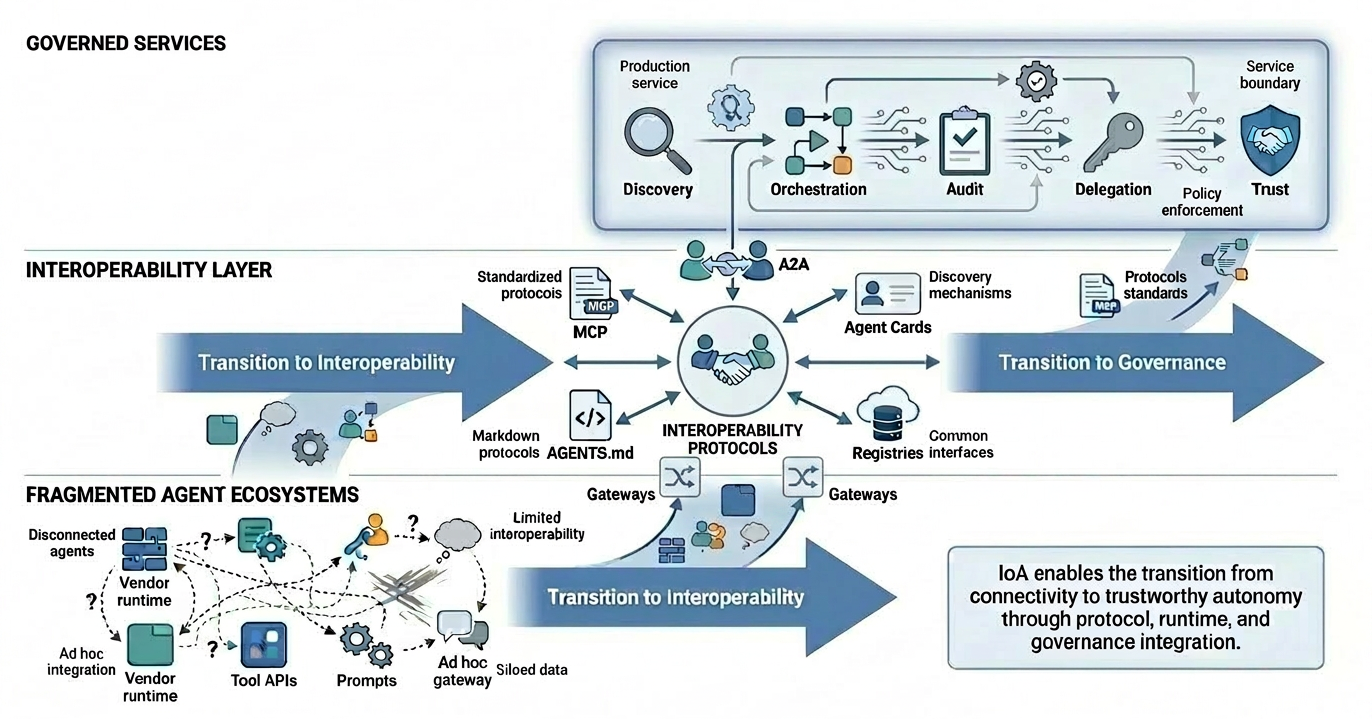}
\caption{The transitions from fragmented agent ecosystems to the Internet of Agentics (IoA) with governance capabilities for discovery, orchestration, audit, delegation, and trust.}
\label{fig:intro}\vspace{-3mm}
\end{figure}

Figure~\ref{fig:intro} illustrates this framing: it depicts the IoA as a transition from fragmented agent ecosystems, in which individual runtimes depend on local tools and ad hoc integration code, toward an explicit interoperability layer composed of protocols, registries, and gateways. The figure further highlights the services-computing implication of this transition: agent systems become viable production services only when interoperability is accompanied by governance mechanisms for discovery, orchestration, auditability, bounded delegation, and trust.

The first wave of LLM deployment favored vertically integrated, provider-specific stacks, whereas the subsequent spread of open-weight models and orchestration frameworks reproduced a familiar pre-standardization pattern: strong local capability coupled with limited cross-system interoperability~\cite{xi2023rise,aaif2025launch,anthropic2024mcp,a2a2025spec}. This pattern motivates the IoA as a services-computing problem rather than merely an agent-engineering problem. The relevant research question is not only whether individual agents can achieve high task performance, but whether independently developed agent systems can cooperate through explicit contracts, observable execution boundaries, and reliability expectations comparable to those already imposed on APIs, workflows, and service platforms.

Reframing agentic interoperability as a challenge of institutionalized service composition, this perspective asserts that basic connectivity is inadequate for autonomous agents that initiate actions, negotiate tasks, and operate across administrative boundaries. Because of this complexity, production-grade ecosystems require robust mechanisms to specify capabilities, mediate coordination, constrain authority, and ensure comprehensive accountability across heterogeneous participants. Consequently, the Internet of Agentics (IoA) aims to transform fragmented agent implementations into a governed, discoverable, and composable cognitive service infrastructure~\cite{papazoglou2003soa,fielding2000rest,a2a2025spec,dean2025delegation} or also namely (ASC) framework proposed by Deng et al.~\cite{asc2025survey}. While the earlier work formalized the agent lifecycle as first-class services encompassing perception, decision-making, and organizational collaboration, we extend this theoretical paradigm by mapping these dimensions onto a concrete protocol-system-service stack to operationalize ASC principles for real-world deployment.

The recent convergence of industry actors around the Agentic AI Foundation (AAIF) gives this architectural transition an institutional basis. MCP standardizes agent access to tools and data sources, A2A specifies peer coordination among opaque agents, and complementary mechanisms such as AGENTS.md, Claude Skills, and secure messaging frameworks begin to provide governance, reusable procedural knowledge, and operational control~\cite{aaif2025launch,aaif2025members,anthropic2024mcp,a2a2025spec,ibm2025acp,agentsmd2025,claudeskills2025,agntcy2025}. Detailed protocol definitions are centralized in Section~\ref{sec:protocols} to avoid repeating them throughout the paper.

The intellectual roots of this development extend beyond contemporary protocol work. Classical multi-agent systems contributed formal agent models, communication languages, and delegation mechanisms, while distributed systems and services computing contributed the interface, workflow, and governance abstractions that modern agentic platforms are now reinterpreting for probabilistic and adaptive components~\cite{rao1995bdi,wooldridge1995intelligent,sycara1998multiagent,finin1994kqml,fipa_acl,fipa2002acl,smith1980cnp,papazoglou2003soa,bpel_webservices}.

\begin{figure}
    \centering
    \includegraphics[width=0.95\linewidth]{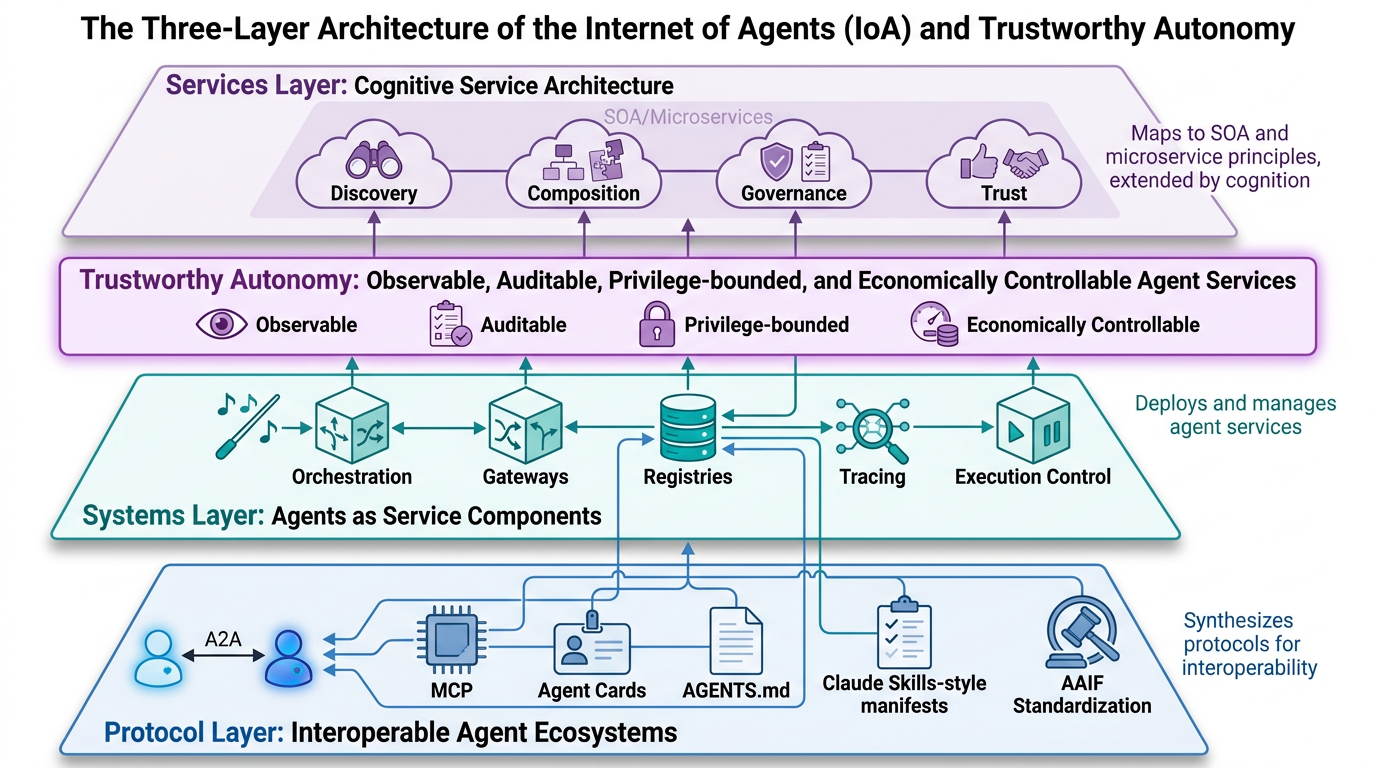}
    \caption{Three-layer architecture of IoA, spanning the protocol, systems, and services layers. The framework captures agent interoperability, deployment infrastructure, and services-computing implications.}
    \label{fig:three-layer}\vspace{-3mm}
\end{figure}

This paper presents an integrative architectural perspective on the emerging Internet of Agentics (IoA), synthesizing protocols, systems, and services-computing developments into a unified framework for trustworthy and governable agentic ecosystems. The paper develops a three-layer architecture of the IoA, as shown in Fig.~\ref{fig:three-layer}, moving from protocols, to system design, to services-computing implications. It makes the following contributions:
\begin{itemize}
	\item It synthesizes the emerging protocol layer for interoperable agent ecosystems, with particular attention to MCP, A2A, Agent Cards, AGENTS.md, Claude Skills-style procedural manifests, and AAIF-led standardization and protocols.
	\item It analyzes the systems layer that turns agents into deployable service components, connecting orchestration, gateways, registries, tracing, and execution control to services-computing concerns.
	\item It maps the services layer of the IoA onto the established lineage of discovery, composition, governance, and trust, showing in which ways agentic cognitive services extend rather than replace earlier SOA and microservice principles.
	\item It identifies trustworthy autonomy as the integrative research agenda for the IoA: agent services must become observable, auditable, privilege-bounded, and economically controllable before they can support production workloads~\cite{dean2025delegation,asc2025survey}.
\end{itemize}
Fig. 2 shows the proposed three-layer view of the Internet of Agentics (IoA), comprising protocol, systems, and services layers, with trustworthy autonomy spanning all layers.

The remainder of the paper follows this progression. Section~\ref{sec:methodology-positioning} explains the perspective methodology, source boundaries, and positioning against closely related works. Section~\ref{sec:protocols} analyzes the emerging protocol stack. Section~\ref{sec:systems} examines logical agent design, orchestration, runtime abstractions, and hardware-aware implementation concerns relevant to deployable agent systems. Section~\ref{sec:services} maps these developments to services computing and representative agentic services. Section~\ref{sec:future} synthesizes the resulting research agenda around trustworthy autonomy, and Section~\ref{sec:conclusion} discusses limitations and concludes.

\section{Methodology and Positioning}
\label{sec:methodology-positioning}

Because much recent Internet of Agentics activity appears first in protocol repositories, foundation documents, industry technical reports, as well as materials from the Linux Foundation, rather than in archival journals~\cite{mcpspec2025,a2a2025spec,aaif2025launch,ioa2025fundamentals,google2025scaling}, we use a structured narrative review rather than a strict PRISMA-style systematic review. The goal is explanatory synthesis: connecting protocol specifications, deployable system abstractions, and service-computing implications in a field whose terminology and governance structures are still stabilizing.

\begin{table*}[t]
\centering
\caption{Comparison with five recent and closely related works.}
\label{tab:survey-comparison}
\scriptsize
\begin{tabularx}{\textwidth}{l >{\raggedright\arraybackslash}X >{\raggedright\arraybackslash}X >{\raggedright\arraybackslash}X >{\raggedright\arraybackslash}X}
\toprule
\textbf{Paper} & \textbf{Primary Focus} & \textbf{Protocol Coverage} & \textbf{Systems Coverage} & \textbf{Services Coverage} \\
\midrule
Masterman et al.~\cite{masterman2024landscape} & Survey of emerging AI-agent architectures & None & Reasoning, planning, tool calling, and agent communication & Limited \\
Guo et al.~\cite{guo2024large} & Survey of LLM-based multi-agent collaboration & Minimal & Collaboration patterns, benchmarks, and challenges & Limited \\
Yan et al.~\cite{communication_llm_mas2025} & Survey of communication-centric LLM-MAS review & Discusses communication protocols conceptually & Communication mechanisms, negotiation, scalability, and security & Limited \\
Mohammadi et al.~\cite{mohammadi2025evaluation} & Survey of evaluation and benchmarking of LLM agents & None & Evaluation objectives, benchmarks, metrics, and tooling & Enterprise evaluation concerns \\
ASC survey~\cite{asc2025survey} & Survey of agentic services computing & Brief MCP/A2A discussion & Focus on software rather than software-hardware synergy & Lifecycle framework and service view \\
\textbf{This paper} & \textbf{Integrative architectural perspective on IoA across protocols, systems, and services} & \textbf{Detailed MCP, A2A, ACP, Agent Cards, AGENTS.md, and AGNTCY/SLIM} & \textbf{Runtime control, orchestration, gateways, registries, and hardware-aware implementation} & \textbf{SOA mapping, domains, governance, and trustworthy autonomy} \\
\bottomrule
\end{tabularx}\vspace{-3mm}
\end{table*}

\subsection{Sources and Scope}

The corpus combines four source types: protocol and standards documents for MCP, A2A, ACP, AGENTS.md, Claude Skills and the Agent Skills specification, and AGNTCY/SLIM~\cite{anthropic2024mcp,mcpspec2025,a2a2025spec,ibm2025acp,agentsmd2025,claudeskills2025,anthropicSkillsBlog2025,anthropicSkillsEngineering2025,agentSkillsSpec2025,skills_vs_mcp2025,agntcy2025}; recent surveys and peer-reviewed work on LLM agents, multi-agent systems, service computing, and agent architectures~\cite{masterman2024landscape,asc2025survey,wang2024survey,guo2024large,communication_llm_mas2025,rao1995bdi,yao2023react,shinn2023reflexion,hong2024metagpt}; industry reports and failure analyses describing production constraints~\cite{ioa2025fundamentals,google2025scaling,mast2025failures}; and foundation governance materials that document AAIF-led standardization and project convergence~\cite{aaif2025launch,aaif2025members}. Sources were collected through citation tracing, targeted searches in Google Scholar, Semantic Scholar, and arXiv, monitoring of relevant protocol repositories and foundation announcements, and review of implementation-oriented technical reports.

Table~\ref{tab:survey-comparison} compares the scope and contributions of this article with those of the five closest recent reviews. Existing work emphasizes agent internals and architectures~\cite{masterman2024landscape,wang2024survey}, multi-agent collaboration and communication~\cite{guo2024large,communication_llm_mas2025}, evaluation~\cite{mohammadi2025evaluation}, or agentic services computing~\cite{asc2025survey}. This work uses that literature as context but focuses on the combined protocol--system--service stack of the IoA. Because the field is changing rapidly through protocol releases, foundation coordination, and production-oriented technical reports~\cite{mcpspec2025,a2a2025spec,aaif2025launch,ioa2025fundamentals,google2025scaling}, the result should be read as an integrative architectural map rather than an exhaustive census of all agentic systems.

\subsection{Methodological Limitations}

This methodology has several limitations. First, a structured narrative review cannot provide the reproducibility guarantees of a full PRISMA-style systematic review. It is appropriate for an emerging architectural field in which specifications, repositories, and foundation announcements often precede archival publication, but it necessarily involves judgment in selecting which systems and standards are most representative~\cite{mcpspec2025,a2a2025spec,aaif2025launch,ioa2025fundamentals}. Second, public documentation is uneven across vendors and communities. Protocol specifications tend to expose interface semantics, whereas production systems often expose only partial details about deployment architecture, performance, governance, or adoption. We therefore treat industry sources as evidence of design intent and ecosystem direction rather than as independently verified performance claims~\cite{ioa2025fundamentals,google2025scaling,mast2025failures}.

Third, the corpus is biased toward English-language and publicly documented materials. Internal enterprise deployments, non-English technical communities, and proprietary agent platforms may therefore be underrepresented. Finally, this paper synthesizes architectural patterns rather than benchmarking implementations under controlled workloads. Claims about protocol maturity, runtime design, service governance, and hardware-aware inference infrastructure should be read as a map of emerging design pressures, not as a definitive ranking of products or platforms~\cite{masterman2024landscape,asc2025survey,ioa2025fundamentals}.

Accordingly, this work should be read not as a purely exhaustive survey, but as an integrative architectural perspective that synthesizes protocols, systems, and services-computing developments into a coherent IoA framework.
\section{Protocols and Standards of the IoA}
\label{sec:protocols}

The first layer in the IoA progression is the protocol layer. It addresses a familiar service-computing problem in a new setting: heterogeneous components must interoperate without pairwise integration work. In the agentic setting, those components include not only services and data sources but also autonomous agents with different internal models, tool sets, procedural skills, and governance boundaries. The resulting standards landscape is best understood as a layered stack rather than as an exclusive standardization contest~\cite{anthropic2024mcp,a2a2025spec,agentsmd2025,claudeskills2025,agntcy2025}.

\subsection{AAIF and Protocol Governance}
\label{sec:aaif}

AAIF was created to prevent the agent ecosystem from fragmenting into incompatible vendor silos~\cite{aaif2025launch,aaif2025members}. Its significance is less the number of member organizations than the fact that protocol governance is being pulled into a neutral institutional setting early in the lifecycle of the ecosystem. For services computing, this institutionalization is consequential because interoperability depends not only on message formats but also on durable processes for versioning, conformance, and change control.

Within this stack, MCP occupies the agent--tool layer, A2A occupies the agent--agent layer, AGENTS.md supplies declarative local governance, Claude Skills package reusable procedural knowledge, and complementary transport and identity work such as AGNTCY and SLIM target discovery, routing, and secure messaging~\cite{anthropic2024mcp,a2a2025spec,agentsmd2025,claudeskills2025,agntcy2025}. The protocols are therefore complementary: they define different boundaries in the same distributed system.

\begin{figure}[t]
\centering
\includegraphics[width=0.95\linewidth]{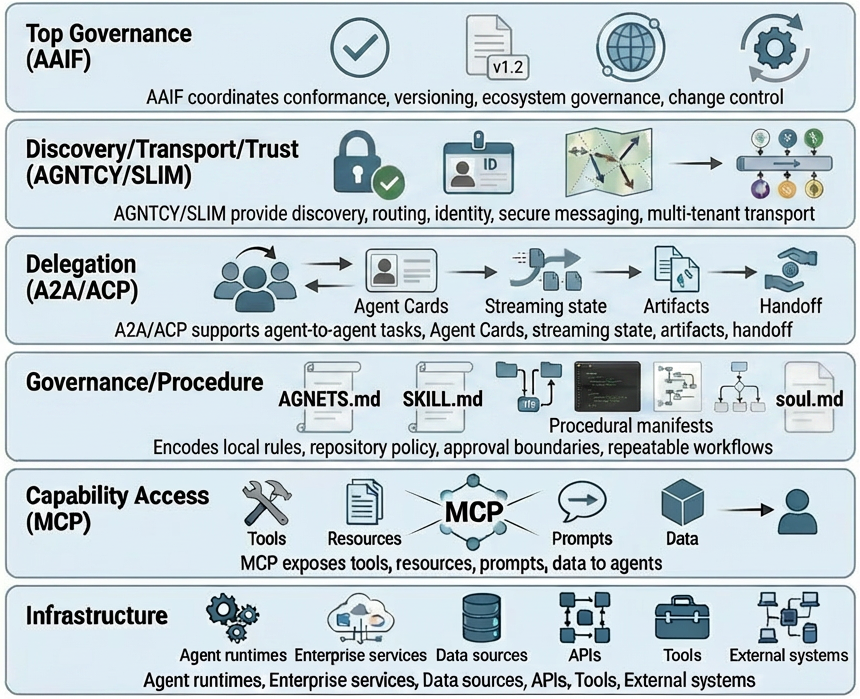}
\caption{Layered IoA protocol stack: agent interoperability from infrastructure and service substrates through MCP-based capability access, A2A/ACP delegation, AGENTS.md governance, Claude Skills-style procedural knowledge, AGNTCY/SLIM discovery and secure transport, and AAIF-led ecosystem governance.}
\label{fig:protocol-stack}\vspace{-3mm}
\end{figure}

Figure~\ref{fig:protocol-stack} summarizes this layered interpretation. It highlights the main organizing principle of the current ecosystem: inter-agent communication, tool access, local governance, procedural skills, and secure transport are being standardized separately because they solve different service-integration problems.

\subsection{Capability Access Layer}
\label{sec:capability-access-layer}

The lowest protocol layer connects agents to tools, data, prompts, and deterministic services. Its design problem is vertical integration: how a reasoning system discovers and invokes external capabilities without a custom adapter for every model--tool pair. MCP is a representative of this layer~\cite{anthropic2024mcp,mcpspec2025,mcp_vs_rest}.

\subsubsection{Model Context Protocol (MCP)}
\label{sec:mcp}

MCP standardizes how an agent discovers and invokes external capabilities~\cite{anthropic2024mcp,mcpspec2025}. Its practical value lies in replacing bespoke adapters for each model--tool pair with a common protocol for exposing tools, resources, and prompts through a stable interface. For services computing, MCP is the main bridge between probabilistic reasoning and deterministic back-end functions, and more concretely a response to the N$\times$M integration problem that earlier API- and plugin-centric ecosystems handled poorly~\cite{anthropic2024mcp,mcp_vs_rest}.

Architecturally, MCP uses JSON-RPC and supports both local and remote transports. The important abstraction is not the wire format itself but the separation of concerns: servers expose capabilities, clients mediate access, and the model decides when to call a tool or retrieve a resource. This reduces integration cost while preserving the possibility of policy control at the host boundary. The client-mediated sampling model is especially important because it prevents servers from directly controlling model outputs and keeps authority with the host application~\cite{anthropic2024mcp,mcpspec2025}.

MCP also highlights the main deployment challenge of agentic services: capability exposure is easy to standardize, but trust is not. Community servers vary in quality, permission models remain inconsistent, and enterprise use still depends heavily on local isolation, auditing, and approval flows. MCP should therefore be read as a necessary interoperability layer, not as a complete governance solution.

This distinction is important for service design. A capability contract can tell an agent what a tool does and how to invoke it, but it does not by itself guarantee that the tool invocation is authorized, safe in context, or aligned with organizational policy. The operational burden therefore shifts toward runtime mediation, which makes MCP most defensible for production use when paired with explicit host-side controls~\cite{anthropic2024mcp,mcpspec2025}.

\subsection{Delegation and Coordination Layer}
\label{sec:delegation-coordination-layer}

Above capability access, the coordination layer treats agents themselves as addressable service actors. Its design problem is horizontal delegation: how one autonomous agent discovers another, negotiates a task, receives progress, and obtains artifacts without requiring shared internals~\cite{a2a2025spec,ibm2025acp}.

\subsubsection{Agent-to-Agent (A2A) Protocol}
\label{sec:a2a}

A2A specifies delegation among autonomous agents~\cite{a2a2025spec}. If MCP answers ``how does an agent call a capability?'', A2A answers ``how does one agent ask another agent to do work?'' The protocol introduces a task model, asynchronous execution semantics, streaming updates, and a discovery mechanism based on Agent Cards. That design is directly relevant to service ecosystems because it makes agents addressable participants in distributed workflows and consolidates prior inter-agent designs such as ACP into a broader industry-facing coordination model~\cite{a2a2025spec,ibm2025acp}.

The key architectural choice in A2A is opaque execution. Agents publish their capabilities and interaction requirements, but they do not have to expose their internal plans, tools, or prompts. This is attractive for enterprise and cross-organizational coordination because it preserves implementation autonomy while still permitting capability-based delegation. At the same time, the same opacity makes debugging, verification, and liability assignment more difficult. In practice, production use will require richer telemetry and audit trails than the protocol alone currently guarantees~\cite{a2a2025spec,wu2023autogen}.

The service-computing implication is that A2A effectively turns agents into discoverable service endpoints with long-running, stateful task semantics. That is materially different from REST-style request/response integration. Task state, streaming updates, human escalation, and artifact return all become part of the service contract. This richer interaction model is useful, but it also raises the bar for observability and conformance~\cite{a2a2025spec,fielding2000rest}.

AAIF's incorporation of ACP into the broader A2A direction reduces fragmentation, but important questions remain open: how Agent Cards should be validated at scale, how multiple bindings should converge operationally, and how coordination failures should be diagnosed when task state alone is insufficient. These are not merely ergonomic issues; they determine whether A2A can support reliable service composition rather than isolated demonstrations~\cite{aaif2025launch,a2a2025spec,ibm2025acp}.

\subsection{Governance and Procedure Layer}
\label{sec:governance-procedure-layer}

Connectivity and delegation are insufficient unless agent behavior is bounded by local policies and repeatable methods. This layer supplies context about what an agent is allowed to do, which conventions it should follow, and how recurring tasks should be executed. Declarative agent manifests such as AGENTS.md and SOUL.md, alongside Claude Skills, are therefore not peripheral prompt files; they are emerging specification-like mechanisms for making agent behavior portable across projects, hosts, and service environments~\cite{agentsmd2025,soulmd2025,claudeskills2025,agentSkillsSpec2025}.

\subsubsection{AGENTS.md Context Provisioning}
\label{sec:agentsmd}

AGENTS.md addresses a different problem than MCP or A2A: local behavioral governance~\cite{agentsmd2025}. It gives agents project-specific instructions about constraints, approval policies, repository conventions, and role boundaries. From a services perspective, it functions as a lightweight governance overlay that constrains how an agent may act once it already has access to tools and collaborators~\cite{agentsmd2025}.

Its importance can be underestimated because it is not a transport protocol. In practice, however, interoperable agent services need local policy hooks that live closer to repositories, workflows, and organizational control points than to network specifications. AGENTS.md is one concrete example of this broader design need~\cite{agentsmd2025}.

\subsubsection{Claude Skills and Agent Skills}
\label{sec:claude-skills}

Claude Skills are reusable procedural packages that describe when an agent should invoke a specialized workflow, what instructions it should follow, and which supporting files, scripts, or assets should be available during execution~\cite{claudeskills2025,anthropicSkillsBlog2025,anthropicSkillsEngineering2025}. They push standardization one layer higher than capability exposure: instead of only exposing tools, they package operational knowledge for how a task should be performed. In service terms, a skill is a portable playbook for agent behavior, analogous to a service recipe or runbook that can be installed, versioned, audited, and reused across task contexts. This distinction matters because dependable agent services require both access and method. Knowing \emph{which} tool exists is not the same as knowing \emph{how} to use it safely and effectively, and analogous abstractions are already appearing across multiple vendor ecosystems~\cite{claudeskills2025,skills_vs_mcp2025,a2a2025spec}.

The broader Agent Skills specification further positions this pattern as an open, cross-platform format rather than only a Claude-specific customization mechanism~\cite{agentSkillsSpec2025,anthropicSkillsSupport2026}. A skill is conventionally organized as a directory with a \texttt{SKILL.md} entry point, YAML front matter, natural-language instructions, and optional supporting files such as scripts, examples, templates, or reference documents~\cite{claudeskills2025,anthropicSkillsRepo2026,anthropicSkillsEngineering2025}. This structure supports progressive disclosure: the agent can see lightweight metadata when selecting a skill, then load the full instructions and supporting material only when needed. From an IoA perspective, that makes skills a procedural standard for packaging local expertise, evaluation lessons, and organizational workflows into reusable agent capabilities.

Table~\ref{tab:mcp-vs-skills} clarifies why this layer is not redundant with MCP. Connectivity protocols and Claude Skills-style procedural manifests solve adjacent but different problems, and mature agent services generally need both.

\begin{table}[t]
\centering
\caption{Connectivity protocols and Claude Skills-style procedural packages as complementary layers in agentic service execution, based on MCP and Skills documentation~\cite{anthropic2024mcp,mcpspec2025,claudeskills2025,agentSkillsSpec2025}.}
\label{tab:mcp-vs-skills}
\small
\begin{tabularx}{\columnwidth}{l >{\raggedright\arraybackslash}X >{\raggedright\arraybackslash}X}
\toprule
\textbf{Dimension} & \textbf{MCP~\cite{anthropic2024mcp,mcpspec2025}} & \textbf{Claude Skills / Agent Skills~\cite{claudeskills2025,agentSkillsSpec2025}} \\
\midrule
Core goal & Capability exposure & Reusable procedural knowledge \\
Mechanism & Protocol interface & \texttt{SKILL.md} packages, Markdown, scripts, and assets \\
Primary value & Reusable tool access & Safer, more consistent method execution \\
Main limit & Weak built-in governance & Weak cross-vendor standardization and validation \\
\bottomrule
\end{tabularx}\vspace{-3mm}
\end{table}

\subsection{Discovery, Transport, and Trust Layer}
\label{sec:discovery-transport-trust-layer}

The outer protocol layer concerns ecosystem-scale operation. Once agents and capabilities span vendors, tenants, and organizations, the stack needs discovery, routing, identity, secure transport, and gateways that can translate among partially overlapping standards~\cite{agntcy2025}.

\subsubsection{AGNTCY and SLIM Messaging}
\label{sec:slim}

AGNTCY and SLIM extend the stack toward discovery, routing, identity, and secure messaging~\cite{agntcy2025}. They matter most when agent ecosystems become multi-tenant and cross-organizational, because at that point capability invocation is only part of the problem; systems also need verifiable identity, secure transport, auditability, and scalable message routing~\cite{agntcy2025}. Their ecosystem visibility and implementation maturity appear less established than MCP or A2A, but they signal where the stack may evolve as production requirements harden.

\subsubsection{Complementary and Emerging Protocols}
\label{sec:other-protocols}

Other efforts, including Agent Protocol, ANP, and IoT-oriented standardization work, illustrate two continuing pressures on the ecosystem~\cite{agentprotocol2024,a2a2025spec}. One is simplification: lightweight protocols remain attractive for local execution and experimentation. The other is decentralization: some use cases demand identity and discovery models that do not rely on centralized registries. These efforts suggest that long-term interoperability may depend on gatewaying and translation rather than on perfect protocol monoculture.

\subsection{Protocol Comparison and Synthesis}
\label{sec:protocol-comparison}

Taken together, the protocol stack suggests a stable conceptual separation. MCP handles vertical integration with tools and data. A2A handles horizontal interaction among autonomous peers. AGENTS.md and procedural manifests constrain local behavior. AGNTCY- and SLIM-like efforts target discovery, trust, and secure transport at larger scale. 

However, with respect to the interoperability goal, this protocol stack appears to provide only the scaffolding for interoperability, while falling short of guaranteeing semantic, behavioral, economic, and governance interoperability required for trustworthy autonomous agent ecosystems. For example, consider an orchestrator delegating a financial risk assessment. \emph{When a sub-agent uses MCP to log a risk entity, the request might succeed syntactically but fail semantically if prompt drift outputs natural language instead of a required enumeration. This issue highlights a critical lack of formal interface models to mathematically align the semantic intent of an agent with a tool's execution boundary}. Furthermore, the ecosystem exhibits immature failure semantics, i.e., \emph{when an agentic workflow halts mid-task, current protocols offer no native state-reconciliation mechanisms--such as automated rollbacks -- to clean up partial database updates}. Additionally, the opaque delegation driven by A2A protocol obscures correctness guarantees; \emph{an orchestrator receiving a  ``Low Risk'' summary cannot currently demand a verifiable provenance graph proving which data was actually analyzed, natively by the protocols}. Ultimately, protocols today standardize connectivity, but the future Internet of Agentics must standardize trust. Until formal models, failure semantics, and correctness guarantees are integrated, trustworthy autonomy will remain heavily reliant on custom runtime infrastructure. 

Table~\ref{tab:protocol-comparison} summarizes the current division of labor across the major protocol families.


\begin{table*}[t]
\centering
\caption{Comparison of major IoA protocol families and the distinct service-integration problem each addresses~\cite{anthropic2024mcp,mcpspec2025,a2a2025spec,agentsmd2025,claudeskills2025,agentSkillsSpec2025,agntcy2025}.}
\label{tab:protocol-comparison}
\small
\begin{tabularx}{\textwidth}{l >{\raggedright\arraybackslash}X >{\raggedright\arraybackslash}X >{\raggedright\arraybackslash}X}
\toprule
\textbf{Protocol Family} & \textbf{Primary Role} & \textbf{Strength} & \textbf{Main Open Issue} \\
\midrule
MCP~\cite{anthropic2024mcp,mcpspec2025} & Agent--tool and agent--data integration & Reduces bespoke adapters through standardized capability exposure & Fine-grained authorization and trust remain runtime-dependent \\
A2A~\cite{a2a2025spec,ibm2025acp} & Agent--agent delegation and coordination & Supports stateful, asynchronous multi-agent workflows & Opaque execution complicates diagnosis, auditing, and liability \\
AGENTS.md / procedural manifests~\cite{agentsmd2025,claudeskills2025,agentSkillsSpec2025} & Local governance and method guidance & Binds agent behavior to project and organizational policy & Lacks a universal cross-vendor operational model \\
AGNTCY / SLIM~\cite{agntcy2025} & Discovery, routing, identity, secure transport & Provides identity, transport, and policy primitives for inter-organization agent meshes & Specifications and conformance suites are still evolving, with limited independent deployment evidence \\
\bottomrule
\end{tabularx}\vspace{-3mm}
\end{table*}

\section{IoA System Design}
\label{sec:systems}

Protocols make interoperability possible, but systems determine whether interoperable agents can be deployed as reliable services. The second layer in the IoA progression is therefore the systems layer: logical agent architectures, orchestration patterns, runtime control planes, and hardware-aware inference infrastructure. From a services-computing perspective, the central systems question is how probabilistic reasoning can be bounded, observed, coordinated, accelerated, and recovered under production conditions. This section therefore treats system design at two connected levels: the logical design of agents and multi-agent runtimes, and the implementation substrate that makes long-running, stateful agent services feasible.

\subsection{Architectural Patterns}
\label{sec:arch-patterns}

The system patterns of the IoA appear at three levels: the internal structure of a single agent, the topology used to coordinate multiple agents, and the control-plane infrastructure used to expose those agents as services~\cite{wu2023autogen,chase2024langgraph}.

\subsubsection{Agent Architectures}
\label{sec:agent-types}

Individual agents range from simple chat interfaces to retrieval-augmented agents, tool-using planners, reflective agents, computer-use agents, and long-horizon coding agents~\cite{yao2023react,shinn2023reflexion}. The main design axes are planning horizon, tool depth, statefulness, and environmental coupling. Systems near the low end of the spectrum remain close to traditional software clients; systems near the high end begin to look like bounded autonomous services with persistent context, approval policies, and execution loops.

From a deployment perspective, the crucial distinction is whether an agent is exposed as an isolated assistant or as a reusable service component. Once exposed as a service, the agent needs explicit contracts around capabilities, memory, execution boundaries, identity, and failure handling. That is the point at which protocols such as MCP and A2A become operational necessities rather than developer conveniences: they make the agent's reasoning loop addressable, delegable, and governable by surrounding software infrastructure~\cite{anthropic2024mcp,a2a2025spec}.

\subsubsection{Multi-Agent Orchestration}
\label{sec:multi-agent-patterns}

Four orchestration patterns recur across current systems. Hierarchical supervision is a common and enterprise-compatible pattern because it simplifies delegation, retries, and result aggregation. Peer-to-peer collaboration appears when teams of agents debate or negotiate without a single controller. Map-reduce styles suit parallelizable analysis, while pipeline styles remain useful when work naturally decomposes into ordered stages~\cite{a2a2025spec,ibm2025acp,hong2024metagpt,wu2023autogen,chase2024langgraph}. These patterns are not novel in distributed systems, but agentic protocols give them a standardized control surface.

Among these patterns, hierarchical supervision appears most compatible with existing service-governance practice because it localizes responsibility and makes oversight tractable. Peer-to-peer collaboration is conceptually appealing but operationally harder to audit. Pipeline and map-reduce compositions are often easier to evaluate because they preserve clearer stage boundaries and output contracts~\cite{wu2023autogen,hong2024metagpt,a2a2025spec}.

\begin{figure}[t]
\centering
\includegraphics[width=0.5\textwidth]{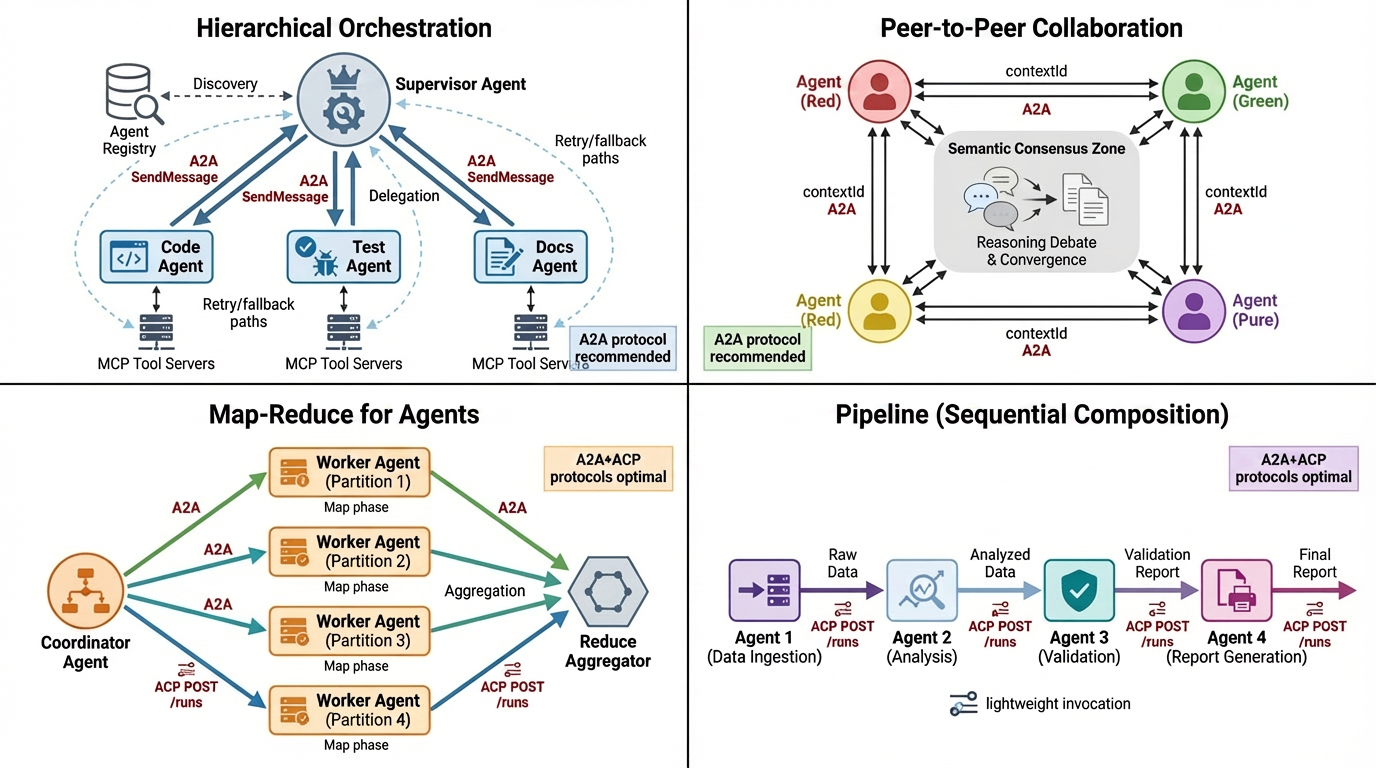}
\caption{Common orchestration patterns in current agent systems. These patterns recur because they balance delegation flexibility against different observability, retry, and policy-enforcement requirements.}
\label{fig:orchestration-patterns}
\end{figure}

Figure~\ref{fig:orchestration-patterns} summarizes these recurring coordination topologies and highlights their different observability burdens. A supervised hierarchy concentrates responsibility and simplifies rollback, whereas peer collaboration complicates provenance. Map-reduce and pipeline forms remain operationally attractive because they preserve stage boundaries and aggregation points.

\subsubsection{Distributed Agentics}
\label{sec:infra-patterns}

At scale, the defining infrastructure components are agent gateways, registries, and runtime managers~\cite{agntcy2025}. Gateways translate protocols, enforce policy, and expose observability. Registries index capabilities and health. Runtimes separate declarative configuration from execution, making agents look more like deployable service units than like ad hoc prompt scripts. This is the systems shift that most clearly links the IoA to mainstream services computing.

The result is a recognizable service architecture. Agent gateways resemble API gateways with added policy and task semantics. Registries resemble service discovery systems with richer capability descriptions. Agent runtimes resemble managed execution environments that bind policy, context, and tool access together. The novelty lies in the probabilistic reasoning core, but the operational scaffolding is still unmistakably that of distributed services~\cite{fielding2000rest,papazoglou2003soa,agntcy2025}.

\subsection{System Implementation and Hardware-Aware Runtime Design}
\label{sec:design-impl}

The implementation layer connects the logical agent patterns above to the physical resources needed to run them at service scale. This subsection separates two implementation questions: how runtime work is divided between deterministic orchestration and model inference, and how inference-serving mechanisms manage long-lived context, latency, and cost.

\subsubsection{CPU--GPU Runtime Bifurcation}
\label{sec:cpu-gpu-runtime}

Agent runtimes increasingly look like control planes that assemble context, call tools, validate outputs, and manage long-lived task state~\cite{welsh2001seda,schmidt2000posa2}. This control plane is software architectural, but it is not hardware-neutral. Agent workloads are dominated by repeated context assembly, LLM inference, tool waiting, validation, and state handoff. As a result, production systems must coordinate two different substrates: CPU-side orchestration for deterministic control and GPU-side inference for probabilistic reasoning.

The CPU side performs event parsing, routing, retrieval, prompt construction, tool execution, policy checks, and validation. The GPU side performs model inference, which is itself split into compute-bound prefill and memory-bound decode phases. This split matters for system design because agent workflows repeatedly alternate between reasoning and external action. While the GPU generates or revises a plan, the CPU can execute tools, collect observations, and prepare the next context. Efficient runtimes therefore resemble staged event-driven systems in which inference, retrieval, tool execution, and validation are pipelined rather than serialized.

\begin{figure}[t]
\centering
\includegraphics[width=0.95\linewidth]{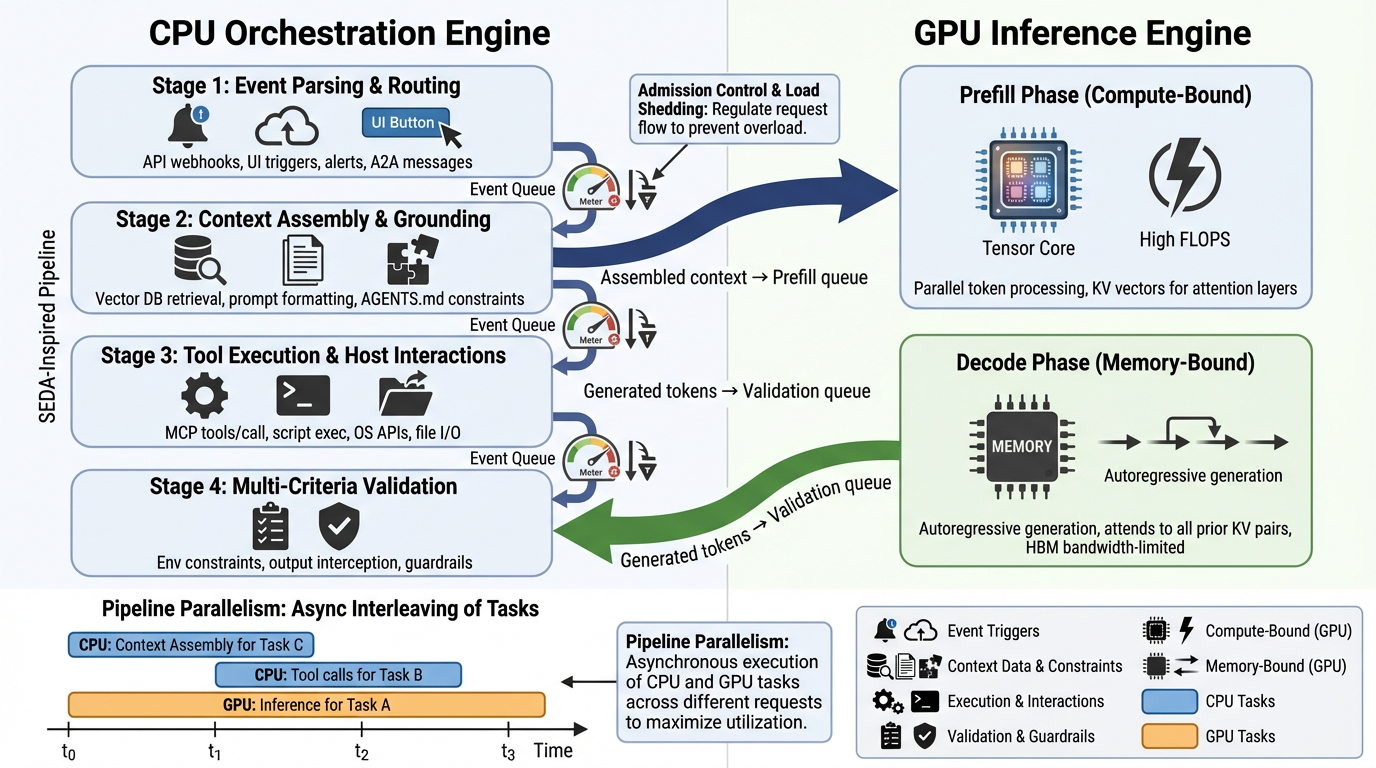}
\caption{CPU--GPU bifurcation in agentic runtime design. CPU-side orchestration handles deterministic routing, context assembly, tool execution, and validation, while GPU-side inference handles model reasoning through prefill and decode phases.}
\label{fig:cpu-gpu-bifurcation}\vspace{-3mm}
\end{figure}

\subsubsection{Inference Serving and Stateful Context}
\label{sec:inference-serving-context}

Hardware-aware inference serving further changes the design space. Prefill--decode disaggregation separates the compute-intensive processing of long prompts from the memory-bandwidth-limited generation of output tokens, allowing different GPU pools or scheduling policies to optimize each phase. Dynamic migration systems such as Llumnix~\cite{sun2024llumnix} and disaggregated serving systems such as DistServe~\cite{zhong2024distserve} are important not merely as accelerator optimizations, but because they make bursty, multi-step agent workloads easier to schedule under latency and cost constraints. For services computing, the practical implication is that the agent runtime must expose budget, latency, and quality controls that reflect the physical cost of reasoning.

The most agent-specific hardware pressure is state management. Long-running workflows repeatedly reuse large contexts, branch into subtasks, and return to earlier evidence. KV cache memory therefore becomes a first-class systems resource rather than an invisible model-inference detail. Techniques such as PagedAttention~\cite{kwon2023vllm}, prefill reuse, and cache-aware scheduling are relevant to agentic services because they affect how quickly delegated tasks can resume and how expensive long-horizon reasoning becomes. We treat these mechanisms as implementation constraints rather than as a separate architectural layer, because the service-facing question is whether the runtime can expose useful latency, cost, and isolation controls to the agent system above it.

These implementation choices do not replace logical agent design; they constrain it. A planner with long context, multiple delegated subtasks, and frequent validation loops imposes different hardware demands from a short-lived retrieval agent. Conversely, hardware-aware runtimes can make richer logical designs practical by reducing redundant prefill, reusing shared context, and hiding latency behind tool execution. The systems layer of the IoA therefore sits between software architecture and physical infrastructure: gateways, registries, and policies make agents governable, while inference serving and memory hierarchy make them operationally feasible. It is important to note, however, that current request-level abstractions do not by themselves capture the complex structure of multi-agent delegation graphs. We detail the physical isolation and context-provenance risks of this gap in Section~\ref{sec:security-hardware-isolation}, and examine how SLA design and cost modeling must evolve to address it in Section~\ref{sec:open-challenges}.

\subsection{Limitations of Current System Architectures}
\label{sec:system-limitations}

The same architectural features that make agent systems useful also create unresolved systems risks. This subsection groups those risks into two categories: coordination and observability failures in the logical runtime, and security or isolation failures that emerge when agents receive persistent authority over tools, context, and infrastructure~\cite{mast2025failures,google2025scaling,a2a2025spec,dean2025delegation}.

\subsubsection{Coordination, Context, and Observability}
\label{sec:coordination-observability-limits}

Current systems remain constrained by four recurring weaknesses. Multi-agent execution still suffers from context degradation and error amplification: long delegation chains either pass forward too much conversational residue or compress intermediate state too aggressively, allowing a small hallucination to propagate downstream~\cite{mast2025failures}. Coordination overhead also grows faster than many demonstrations suggest; gains are strongest on naturally parallel workloads, whereas sequential workflows often accumulate latency, token cost, and verification overhead faster than they gain quality~\cite{google2025scaling,mast2025failures}.

Observability and accountability are likewise immature relative to conventional service platforms. Opaque execution is often desirable at the protocol boundary, but today's runtimes still expose limited diagnostics about why a delegated task failed, which context was used, how tools were selected, or where policy violations emerged~\cite{a2a2025spec}. Existing coding-agent frameworks make progress through validation loops and approval checkpoints, yet the broader ecosystem still lacks stable norms for execution tracing, conformance testing, and post hoc audit.

\subsubsection{Security and Hardware Isolation}
\label{sec:security-hardware-isolation}

Security and privilege management are also not commensurate with the authority granted to modern agents. Application-layer prompts and allowlists are insufficient once agents can act persistently on external systems or delegate work to other agents. More mature systems will need bounded credentials, stronger sandboxing, revocable delegation, and clearer separation between reasoning authority and execution authority~\cite{a2a2025spec,birgisson2014macaroons,dean2025delegation}.

From a service-governance perspective, hardware-aware implementation still exposes unresolved maturity problems. Disaggregated serving, cache sharing, and memory pooling improve feasibility, but they introduce new scheduling and isolation problems. Shared context caches can reduce cost, yet they also raise questions about tenant separation, cache provenance, and side-channel leakage. Similarly, moving state across GPU, CPU, CXL, NVMe, and DPU-managed network tiers can make reasoning cheaper, but it makes reproducibility and fault diagnosis harder unless runtimes preserve clear execution traces. Mature IoA systems therefore need diagnostics that connect logical events---prompts, retrieved context, tool calls, approvals, and side effects---to physical events such as cache reuse, migration, inference latency, and resource consumption.

\section{Agentic Services Computing}
\label{sec:services}

With the protocol and system layers in place, the third layer in the IoA progression is the services layer. The main question for services computing is what agents contribute beyond conventional service composition. Our argument is that agentic systems do not replace service-oriented thinking; they relocate it. The system boundary shifts from deterministic functions toward protocol-described agent and tool capabilities, but the core service concerns remain discovery, orchestration, policy, and trust~\cite{papazoglou2003soa,bpel_webservices,a2a2025spec}.

\subsection{ASC Framework}
\label{sec:asc-framework}

The agentic services computing framework (the term was firstly coincided in~\cite{asc2025survey}) explains how conventional service-engineering concepts reappear when the service actor is no longer only a deterministic endpoint but a bounded reasoning system. Here we extends the ASC framework through three lenses: the historical recurrence from SOA to agentic services, the changing role of data and semantics, and the lifecycle implications of deploying agents as governed services~\cite{papazoglou2003soa,fielding2000rest}.

\subsubsection{Cyclical Thesis}
\label{sec:cyclical-thesis}

The clearest services-computing interpretation of the IoA is one of architectural recurrence rather than wholesale replacement~\cite{papazoglou2003soa}. RPC becomes model-mediated function invocation. BPEL-style orchestration becomes agent planning and delegation. SOAP/WSDL/UDDI-style interface and discovery concerns reappear in MCP and Agent Cards. Microservice boundaries become capability boundaries, where an agent may serve as an intelligent facade over conventional services while deterministic microservices continue to handle precise operations. The ESB reappears in more decentralized form through agent coordination protocols such as A2A~\cite{bpel_webservices,uddi_mcp_comparison,a2a2025spec}.

This analogy is useful but incomplete. Agentic systems add failure modes that traditional SOA did not face, including hallucination cascades, context degradation, and non-deterministic coordination. The value of the cyclical thesis is therefore diagnostic, not predictive~\cite{mast2025failures,google2025scaling}.

As formalized in Table~\ref{tab:soa_agentic_mapping}, the transition from SOA to the IOA represents an evolutionary relocation of core services computing principles. While classical services computing systems relied on deterministic, structural coupling via RPC and microservice interfaces, the agentics introduces a paradigm shift toward semantic, intent-based late binding and cognitive capability boundaries. Similarly, the static finite state machines that once dictated BPEL/BPMN workflow orchestration are supplanted by emergent, probabilistic cognitive loops. The point is not that old paradigms can simply be renamed; rather, the same fundamental design pressures--such as remote invocation, integration middleware, and state management--recur at a higher layer of abstraction. Ultimately, while services computing once standardized deterministic software interfaces, the IoA reinterprets these boundaries to govern the interaction among bounded reasoning systems.  



\begin{table*}[t]
\centering
\caption{Contrasting SOA with Agentic AI: A paradigm shift from deterministic, programmatic composition to autonomous, LLM-based orchestration.}
\label{tab:soa_agentic_mapping}
\small
\begin{tabularx}{\textwidth}{l >{\raggedright\arraybackslash}X >{\raggedright\arraybackslash}X >{\raggedright\arraybackslash}X}
\toprule
\textbf{Dimension} & \textbf{Historical Paradigm} & \textbf{Agentic Equivalent} & \textbf{Key Architectural Shift} \\
\midrule
Remote Invocation & RPC / gRPC / CORBA & LLM Function Calling & Deterministic coupling $\rightarrow$ semantic intent-based late binding \\
Workflow Orchestration & BPEL / BPMN & Agentic Workflows (ReAct, CoT) & Static finite state machines $\rightarrow$ emergent cognitive loops \\
Interface / Discovery & SOAP / WSDL / UDDI & MCP (JSON-RPC manifests) & Manual WSDL integration $\rightarrow$ LLM-driven autonomous discovery \\
Service Decomposition & REST Microservices & Agentic Microservices & Business logic boundaries $\rightarrow$ cognitive capability boundaries \\
Integration Middleware & Enterprise Service Bus & A2A Protocol & Centralized hub-and-spoke $\rightarrow$ decentralized peer-to-peer \\
Data Representation & XML, JSON (syntactic) & Vector Embeddings (semantic) & Keyword matching $\rightarrow$ geometric similarity in latent space \\
State Management & Stateless REST & Deeply Stateful (KV Cache, Memory) & Ephemeral requests $\rightarrow$ persistent context across multi-step workflows \\
\bottomrule
\end{tabularx}
\footnotesize{\textit{Note:} The mappings shown are conceptual analogies intended to highlight shifts in abstraction and system design. They do not imply strict architectural equivalence.}\vspace{-3mm}
\end{table*}








\subsubsection{Data Standards and Semantics}
\label{sec:data-standards}

Agentic services also alter data handling. XML and JSON remain essential interchange formats, but they are increasingly complemented by vector representations that support semantic retrieval and cross-modal matching~\cite{lewis2020rag}. This does not eliminate structured data management; rather, it adds a semantic access layer above conventional service contracts. For service composition, the practical implication is that retrieval quality and grounding become part of the reliability story, especially as hybrid relational-vector systems become part of mainstream service infrastructure~\cite{lewis2020rag,dean2025sequoia,facts2025grounding}.

This observation is central because service reliability now depends partly on information selection before execution. If an agent service retrieves the wrong context, downstream reasoning can fail even when the surrounding infrastructure is correct. Grounding, retrieval quality, and evidence ranking therefore become service-quality concerns rather than only model-quality concerns~\cite{lewis2020rag,facts2025grounding}.

\subsubsection{ASC Lifecycle}
\label{sec:asc-lifecycle}

Agentic services also expand the service lifecycle. Design now includes capability scoping and policy definition. Deployment includes runtime configuration and model binding. Operation includes iterative reasoning, tool use, and cross-agent coordination. Evolution includes agent-capability updates, prompt, policy, workflow, and skill refinement over time~\cite{yao2023react,shinn2023reflexion,claudeskills2025}. Claude Skills make this evolution concrete: recurring failures, new tools, domain conventions, or improved procedures can be folded back into reusable skill packages, creating a controlled self-evolution loop in which the agent service improves its own procedural substrate while remaining subject to review, versioning, and regression tests. Trustworthiness cuts across all four phases, which is why services governance becomes central rather than peripheral. The lifecycle lens is useful because it shows that agentic services add managerial work at every stage rather than only at runtime.

Table~\ref{tab:asc-lifecycle} makes this lifecycle shift explicit by mapping each phase to its service purpose and corresponding protocol--system design concerns. The table highlights that security, evaluation, value alignment, and policy control cannot be postponed until deployment; they shape design, runtime operation, and post-deployment evolution as continuous service-management responsibilities.

\begin{table}[t]
\centering
\caption{Agentic services computing lifecycle across service purpose and design responsibilities, synthesized from agentic services computing, agent runtime, protocol, and procedural-skill sources~\cite{yao2023react,shinn2023reflexion,anthropic2024mcp,a2a2025spec,agentsmd2025,claudeskills2025}.}
\label{tab:asc-lifecycle}
\scriptsize
\renewcommand{\arraystretch}{1.15}
\begin{tabularx}{\columnwidth}{>{\raggedright\arraybackslash}p{0.18\columnwidth} >{\raggedright\arraybackslash}p{0.28\columnwidth} >{\raggedright\arraybackslash}X}
\toprule
\textbf{Phase} & \textbf{Purpose} & \textbf{Design} \\
\midrule
Design & Define the agentic service contract: task intent, QoS/SLA targets, risk class, approval gates, and value metric. & Translate requirements into \emph{systems} architecture and \emph{services} interfaces: MCP tools, A2A boundaries, Agent Cards, AGENTS.md constraints, memory, orchestration, validation, and supervision~\cite{anthropic2024mcp,a2a2025spec,agentsmd2025}. \\
Deployment & Instantiate the agentic service for a tenant: permissions, compliance policy, data boundaries, SLA tier, and billing model~\cite{dean2025delegation}. & Bind \emph{systems} endpoints, schemas, credentials, identity, discovery, and conformance assumptions; configure \emph{services} routing, retrieval, sandboxing, observability, budgets, and isolation~\cite{a2a2025spec,agntcy2025}. \\
Operation & Run the agentic service under QoS/SLA commitments: bounded delegation, escalation, audit, accountability, and value delivery~\cite{dean2025delegation}. & Coordinate \emph{systems} state, tool results, artifacts, handoffs, and policy signals across MCP/A2A interfaces; manage \emph{services} traces, validation, scheduling, latency, cost, and recovery~\cite{anthropic2024mcp,a2a2025spec}. \\
Evolution & Evolve the agentic service contract: update agent capabilities, pricing, compliance evidence, provenance, and trust controls. & Version \emph{systems} protocols, capability descriptions, Agent Cards, governance files, and Claude Skills; refine \emph{services} prompts, workflows, caches, evaluation harnesses, and regression tests from incidents~\cite{claudeskills2025,anthropicSkillsEngineering2025,agentSkillsSpec2025}. \\
\bottomrule
\end{tabularx}\vspace{-3mm}
\end{table}

\subsection{Service Domains and Web 4.0}
\label{sec:service-domains}

The clearest early service domains are those already rich in orchestration and policy. Telecommunications, enterprise workflow automation, software engineering, scientific pipelines, and financial services all fit this pattern because they require coordination across heterogeneous capabilities under explicit constraints~\cite{a2a2025spec}. In these domains, agents are valuable not because they replace deterministic services, but because they sit above them as adaptive coordinators. A credible near-term deployment pattern is therefore hybrid: agents orchestrate reliable service components rather than attempting to subsume them. The following discussion moves from concrete representative services to broader domain patterns, so the examples first show service boundaries in practice and the summary table then abstracts the recurring roles, backbones, and governance pressures.

\subsubsection{Representative Agentic Services}
\label{sec:rep-systems}

Representative deployments clarify this services view. Claude Code emphasizes composable developer workflow integration, while Goose emphasizes local-first execution and explicit control over context and tools~\cite{claudecode2025,goose2025arch}. Enterprise copilots and AI coworkers such as Microsoft 365 Copilot, Claude CoWork, and Gemini for Google Workspace illustrate a different service pattern: agentic assistance embedded inside existing productivity and collaboration suites, grounded in organization-specific content, and constrained by identity, permissions, compliance controls, and administrative governance~\cite{m365copilot_arch2026,m365copilot_privacy2026,microsoftgraph_overview2026,anthropic_claude_work2026,google_gemini_workspace2026}. These examples belong in the services layer because their defining contribution is not a new protocol or inference algorithm, but a service boundary that connects reasoning to organizational work.

Table~\ref{tab:system-archetypes} summarizes the service-runtime implication that each representative deployment foregrounds. The table is not intended as a product comparison; it identifies recurring design pressures, from supervised validation in coding workflows to identity-grounded governance in enterprise productivity environments.

\begin{table*}[t]
\centering
\caption{Representative agentic services and the service-runtime tradeoffs they foreground~\cite{claudecode2025,goose2025arch,m365copilot_arch2026,m365copilot_privacy2026,microsoftgraph_overview2026,anthropic_claude_work2026}.}
\label{tab:system-archetypes}
\footnotesize
\begin{tabularx}{\textwidth}{l >{\raggedright\arraybackslash}X >{\raggedright\arraybackslash}X}
\toprule
\textbf{Service} & \textbf{Primary Emphasis} & \textbf{Key Service Implication} \\
\midrule
Claude Code~\cite{claudecode2025} & Composable developer workflow integration & Strong fit for supervised orchestration and iterative validation \\
Goose~\cite{goose2025arch} & Local-first execution and explicit control & Strong fit for privacy-sensitive deployments with bounded tool access \\
Microsoft 365 Copilot~\cite{m365copilot_arch2026,m365copilot_privacy2026,microsoftgraph_overview2026,m365copilot_connectors2026} & Enterprise productivity assistance grounded in Microsoft Graph & Highlights identity-scoped grounding, tenant data boundaries, Purview compliance, auditability, and admin-governed extensibility \\
Claude CoWork~\cite{anthropic_claude_work2026} & Enterprise AI coworker for organizational knowledge and shared projects & Highlights collaborative reasoning over governed work context, artifacts, projects, and workspace administration \\
\bottomrule
\end{tabularx}\vspace{-3mm}
\end{table*}

\begin{figure}[t]
\centering
\includegraphics[width=0.5\textwidth]{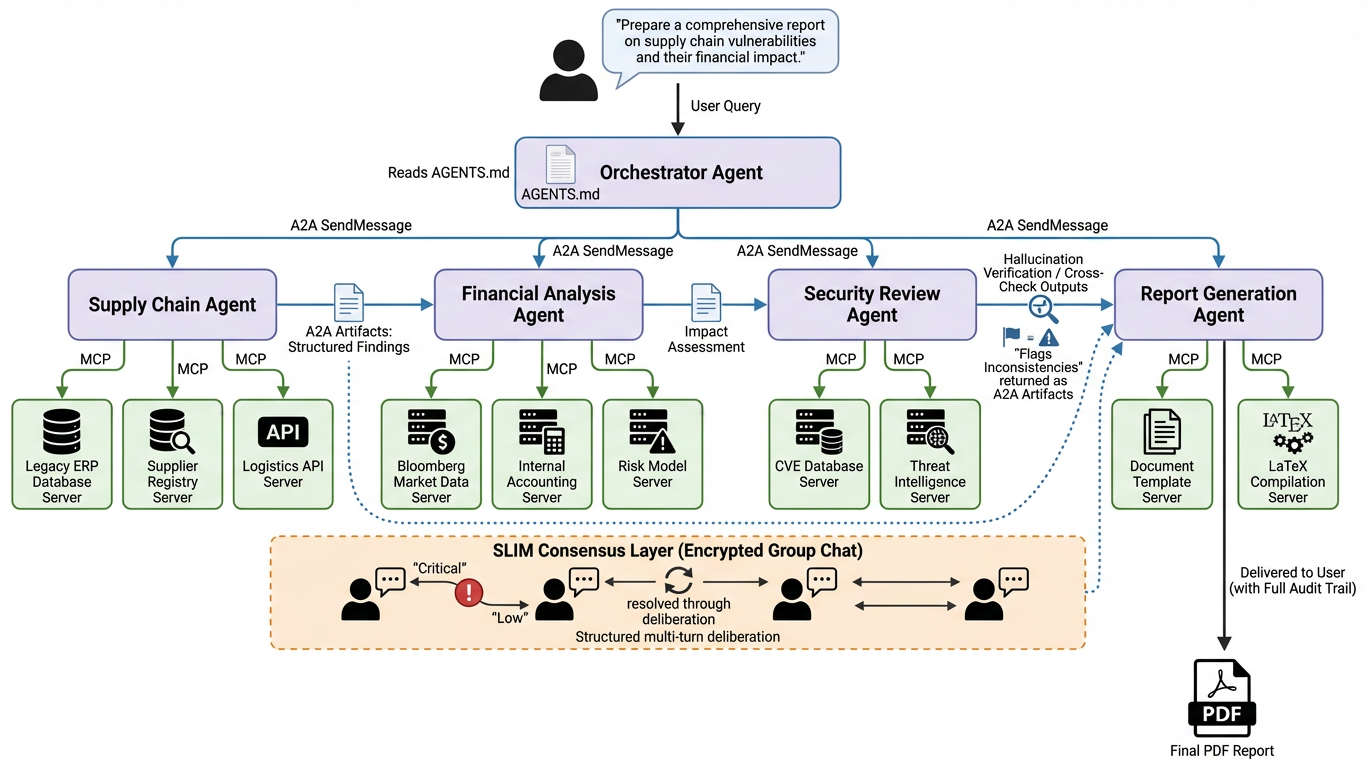}
\caption{Enterprise agentic mesh workflow illustrating how an orchestrator decomposes a high-level request into specialized services connected through protocol-governed delegation and tool access.}\vspace{-3mm}
\label{fig:enterprise-mesh}
\end{figure}

Figure~\ref{fig:enterprise-mesh} illustrates this hybrid deployment pattern. A high-level request is decomposed by an orchestrating agent into delegated subtasks, while deterministic back-end services remain responsible for data access, validation, execution, and record updates. This separation is important because it preserves the reliability of conventional service components while allowing agents to coordinate across them.

\subsubsection{Enterprise and Finance}

Enterprise service domains are attractive for agentic deployment because they already contain explicit process boundaries, policy requirements, and heterogeneous back-end systems. In supply-chain analysis, compliance review, or financial reporting, the main difficulty is often not computation itself but coordination across multiple data sources and service interfaces. An agentic layer can reduce the amount of custom integration logic by translating a high-level request into a set of structured subtasks, invoking the required services, and reconciling intermediate results before producing a final artifact~\cite{a2a2025spec}.

Microsoft 365 Copilot illustrates this enterprise-service pattern in a production productivity environment. Rather than presenting an agent as a free-standing reasoning endpoint, it embeds agentic assistance into Microsoft 365 applications, grounds responses through Microsoft Graph, and constrains access through tenant permissions and organizational controls~\cite{m365copilot_arch2026,m365copilot_privacy2026,microsoftgraph_overview2026}. Its extension model also shows how agentic services can connect to external enterprise knowledge: Copilot connectors bring line-of-business content into Graph-based retrieval or federated access patterns while preserving admin consent, user permissions, and connector governance~\cite{m365copilot_connectors2026}. Claude CoWork and Gemini for Google Workspace follow the same broad pattern in different collaboration suites: the agent is valuable because it operates over governed organizational context rather than as an isolated model endpoint~\cite{anthropic_claude_work2026,google_gemini_workspace2026}. These systems make enterprise agentic services computing concrete: the agent layer coordinates across documents, communications, meetings, and external systems, while deterministic identity, compliance, search, and data-governance services bound its behavior.

Financial services provide an even stricter version of this requirement. Market interpretation, strategy synthesis, and anomaly triage are all plausible agentic tasks, but deterministic services remain essential for trade execution, portfolio accounting, regulatory checks, and record updates. This makes finance a useful model for the broader IoA: the future is most likely to be hybrid, with agentic reasoning layered above reliable transactional services rather than replacing them outright.

\subsubsection{Scientific, Engineering, and Local Automation}

Scientific workflows and software-engineering pipelines also fit the IoA pattern because they require iterative reasoning over diverse tools rather than a single monolithic API call. In research review, literature triage, or engineering validation, an agentic service can gather evidence, call specialist tools, compare outputs, and assemble a structured conclusion while leaving consequential decisions subject to human verification~\cite{a2a2025spec}. Coding agents and research-oriented agents are useful examples for the same reason: their value comes from orchestrating already useful service components under a higher-level reasoning loop. Claude Code and Goose make this pattern visible in developer services. Both expose the full loop---repository retrieval, plan formation, tool invocation, validation, and iterative repair---but place the boundary differently: Claude Code emphasizes composable workflow integration and supervised validation, whereas Goose emphasizes local-first execution for privacy-sensitive settings~\cite{claudecode2025,goose2025arch}.

Table~\ref{tab:service-domains} summarizes the cross-domain pattern after these examples. It shows where deterministic substrates remain indispensable and where governance pressure is strongest across enterprise, finance, scientific, engineering, and local automation settings.

\begin{table*}[t]
\centering
\caption{Representative service domains for the IoA and the hybrid architectures they favor, synthesized from agentic-services, protocol, enterprise-copilot, and developer-agent sources~\cite{a2a2025spec,m365copilot_arch2026,claudecode2025,goose2025arch}.}
\label{tab:service-domains}
\footnotesize
\begin{tabularx}{\textwidth}{l >{\raggedright\arraybackslash}X >{\raggedright\arraybackslash}X >{\raggedright\arraybackslash}X}
\toprule
\textbf{Domain} & \textbf{Agentic Role} & \textbf{Deterministic Service Backbone} & \textbf{Primary Governance Pressure} \\
\midrule
Enterprise workflows~\cite{m365copilot_arch2026,m365copilot_privacy2026,microsoftgraph_overview2026,m365copilot_connectors2026} & Intent translation, knowledge grounding, and exception handling & ERP, Microsoft Graph, document, identity, reporting services & Auditability, policy conformance, and data-governance controls \\
Financial services & Interpretation, triage, analysis synthesis & Execution, accounting, compliance services & Risk controls and regulatory traceability \\
Scientific and engineering workflows~\cite{a2a2025spec} & Evidence gathering and tool chaining & Databases, simulation, build-validation services & Reproducibility and provenance \\
Developer and local automation~\cite{claudecode2025,goose2025arch} & Repository assistance and local orchestration & File systems, build tools, terminals, OS sandboxes, and schedulers & Containment, privilege protection, and reproducible validation \\
\bottomrule
\end{tabularx}\vspace{-3mm}
\end{table*}

\subsubsection{The Agentic Web}
\label{sec:web4}

The idea of a ``Web 4.0'' or agentic web is useful as a directional metaphor, but it should be treated cautiously. The strongest near-term interpretation is not fully autonomous machine commerce; it is a web in which agents become first-class service consumers and providers under explicit governance. If that vision matures, the protocols surveyed here may play a role for agent interoperability analogous to what earlier web standards did for documents and APIs~\cite{mcp_a2a_tcpip}. Whether that happens depends less on visionary framing than on conformance, security, and deployment economics.

\subsection{Agentics or Services: an Open Debate}
\label{sec:agentics-versus-services-subscription}

The emergence of agentic systems raises a fundamental question for services computing: should agents be treated as applications consumed by users, or as first-class service actors that compose, invoke, and deliver functionality within larger service ecosystems and workflows? The question is not mereley semantics.  It reflects a deeper shift in the unit of computation, the unit of value, and the nature of service abstraction~\cite{chaisiri2012resourcecost,chatterjee2017dynamicpricing,saxena2024oversubscriptionpricing}.

Unlike traditional software services, which expose deterministic functionality with relatively stable cost and performance characteristics, agentic systems exhibit dynamic execution behavior, uncertain outcomes, and variable resource consumption. As a result, treating agents as conventional services risks obscuring critical properties such as cost, latency, and confidence, which are essential for reliable operation at scale. This section examines the implications of this shift, focusing on how agentic execution redefines service composition, economic models, and system-level guarantees.

\subsubsection{Service-Economic Unit Shift}
\label{sec:service-economic-unit-shift}

Classical services computing grew with subscription economics, where the unit of value was access by an authorized human or organization. SOA, service registries, orchestration engines, access-control systems, and SLA management therefore assumed relatively stable subscribers, interfaces, and entitlement boundaries~\cite{papazoglou2003soa,bpel_webservices}. Prior services-computing research has nevertheless long treated services as economic objects, including cost-optimized cloud provisioning, game-theoretic service pricing, dynamic pricing for heterogeneous sensor-cloud services, security-service market investment, and profit-aware resource oversubscription~\cite{chaisiri2012resourcecost,liu2016servicepricingcps,chatterjee2017dynamicpricing,feng2022cloudsecuritypricing,saxena2024oversubscriptionpricing}. 

Agentics extends this line of work because an agentic service is not a single endpoint call. It is a bundle of model inference, tool execution, validation, and sometimes human review. To make this economic shift precise on the major costs--especially \emph{token consumption} and \emph{latency} (both caused by the use of LLMs)--we model execution as a probabilistic directed graph, or namely a random graph, $G = (V, E)$. Here, nodes represent heterogeneous operations including LLM inference, tool invocation, validation, and human interaction. 

Unlike traditional static graphs, this graph is not known \textit{a priori}. Because LLMs sample from probability distributions, the graph is constructed incrementally, evolving randomly through emergent branching, delegation, and retries. Each node $v \in V$ incurs an expected cost $\mathbb{E}[C(v)]$ (e.g., token pricing and API fees) and an expected latency $\mathbb{E}[L(v)]$, and produces an output with a success probability $Q(v) \in [0,1]$. This model enables defining the expected \textit{Economic Quality of Service} (E-QoS) for an execution graph as the tuple:
\begin{equation}
E(G) = \big<\mathbb{E}[C(G)], \mathbb{E}[L(G)], Q(G)\big>,
\end{equation}
where $\mathbb{E}[C(G)] = \sum_{v \in V} \mathbb{E}[C(v)]$ denotes the expected total cost and $\mathbb{E}[L(G)]$ denotes the expected critical-path latency. 

Modeling the workflow as a probabilistic graph exposes fundamental economic risks that deterministic SOA models ignore. Most notably, ensuring an agent completes a complex task without exhausting its budget becomes a \textit{stochastic reachability problem} \cite{puterman2014markov}. This problem introduces significant operational and economic complexity, as probabilistic execution risks massive \emph{sunk token costs from non-terminating or prematurely halted reasoning loops}, alongside severe \emph{path sub-optimality where even successful tasks take unnecessarily expensive, heuristic-driven routes}.

    


Consequently, agent orchestration cannot simply minimize a static graph topology. It must be formulated as a policy optimization problem under uncertainty—minimizing expected cost while providing formal guarantees that latency and quality constraints are met with high probability \cite{hwang2014service}:
\begin{equation}
\begin{aligned}
    \min_{\pi} \quad \mathbb{E}[C(G_\pi)] \quad 
    \text{s.t.}  \quad & P(L(G_\pi) \leq L_{\mathrm{SLA}}) \geq 1-\delta, \\ & \mathbb{E}[Q(G_\pi)] \geq Q_{\mathrm{SLA}}
\end{aligned}
\end{equation}
where $\pi$ represents the orchestration policy governing dynamic graph expansion, and $\delta$ bounds the maximum allowable risk of SLA violation, while $Q_{\mathrm{SLA}}$ and $L_{\mathrm{SLA}}$ refers to the minimal success rate and maximal latency, expected by SLA, respectively. This perspective shifts agentic systems from heuristic orchestration toward mathematically verifiable, risk-aware service composition.

This shift produces a profound economic dilemma: if the value unit is a completed workflow, how should platforms charge for agentic services? Traditional billing mechanisms are fundamentally misaligned with probabilistic execution. While raw LLM access is easily metered by tokens, charging complex agentic workflows per-token forces the consumer to bear the financial risk of non-completion or path sub-optimality, as heavily consumed tokens do not guarantee a desired outcome. Conversely, seat-based subscription models expose service providers to unbounded inference costs from heavy utilization, while unfairly penalizing infrequent users. An outcome-driven pricing model appears to be the logical alternative, yet it remains practically intractable because LLM-driven agents currently lack the strict correctness and success guarantees of classical deterministic services computing. Industry and investment analyses increasingly frame this tension as a question of whether agentic AI could ``eat software''\footnote{``Will AI Eat Software?'' \url{https://www.goldmansachs.com/insights/top-of-mind/will-ai-eat-software}.}, earn durable infrastructure returns\footnote{``Gen AI: Too Much Spend, Too Little Benefit?'' \url{https://www.goldmansachs.com/insights/top-of-mind/gen-ai-too-much-spend-too-little-benefit}.}, and generate enough end-user revenue to support the build-out\footnote{``AI's \$600B Question'' \url{https://www.sequoiacap.com/article/ais-600b-question/}.}.

This shift also exposes a deeper tension in service abstraction. A useful perspective emerges when comparing agentic services to earlier efforts such as Service-Oriented Device Architectures (SODA)~\cite{chen2009soda,deugd2006soda}. In SODA, inherently fragile and resource-constrained devices were abstracted as stable services, often over-promising reliability and safety while masking underlying limitations. This abstraction mismatch led to well-documented challenges in smart environments and cyber-physical systems, including unreliable sensing, intermittent connectivity, and safety risks arising from exposing physical devices as easily invokable services~\cite{helal2020monkey}. In contrast, agentic systems invert this tension: powerful, computationally expensive, and dynamically evolving agents are exposed through simplified service interfaces that under-represent their cost, non-determinism, and operational complexity. This observation reveals a broader duality in service abstraction: when critical properties of the underlying entity are suppressed, whether fragility in devices or expansiveness in agents, system-level risks emerge. In the former case, reliability and safety degrade; in the latter, cost, predictability, and control become the dominant challenges. This duality suggests that future agentic service models must explicitly surface execution bounds, economic cost, and uncertainty as first-class service properties, rather than treating them as hidden implementation details.

\subsubsection{Transaction Costs in Human-Agent Service Work}
\label{sec:services-debate-h2h}

Agentic service execution reduces certain forms of coordination overhead while introducing new sources of system complexity. In traditional workflows, task decomposition, information retrieval, and synthesis often require explicit human coordination across multiple tools and services. Agentic systems can partially automate these steps by constructing and executing an Agentic Service Graph (ASG), thereby reducing the need for manual orchestration~\cite{malone1987coordination,bernstein2012crowdsourcing}.

However, this reduction in human coordination shifts complexity into the execution graph itself. The dynamic expansion of the ASG introduces variability in cost, latency, and outcome quality, making it difficult to predict resource consumption or guarantee performance. Similar challenges have been observed in multi-agent systems and workflow orchestration, where coordination overhead and uncertainty increase with system scale and autonomy~\cite{wooldridge2009introduction,van2003workflow}.

Furthermore, as agents take on greater responsibility for task decomposition and execution, new forms of oversight become necessary, including validation checkpoints, audit trails, and human-in-the-loop escalation mechanisms~\cite{amershi2019guidelines}. From a services perspective, the key challenge is not simply reducing transaction cost, but redistributing it: human coordination costs are replaced by system-level costs associated with monitoring, validation, and control of agentic execution. This shift reinforces the need for explicit modeling of execution graphs and their associated economic and operational properties.

\subsubsection{Metering, Contracting, and Market Design}
\label{sec:services-subscription-model-forecast}

The shift from user-centric applications to agentic service execution necessitates a corresponding shift in how services are metered, priced, and contracted. In traditional service models, billing is typically tied to user access (e.g., subscription or seat-based pricing) or to individual API calls. In contrast, agentic systems execute composite workflows represented as Agentic Service Graphs (ASGs), where a single user request may trigger a sequence of heterogeneous operations including model inference, retrieval, tool invocation, and validation~\cite{cardellini2011qos}.

This compositional nature of execution implies that the economically relevant unit is no longer a single service invocation, but the entire execution graph. Consequently, pricing and metering mechanisms must account for the aggregate cost, latency, and quality characteristics of the ASG. This suggests a transition toward outcome-oriented or graph-based pricing models, where charges reflect the resources consumed and the guarantees provided across the full execution path~\cite{buyya2009market,calheiros2015workload}.

From a systems perspective, this raises several challenges. First, the dynamic structure of ASGs makes cost estimation non-trivial prior to execution. Second, variability in execution paths complicates the enforcement of service-level agreements (SLAs), particularly when quality and confidence must be considered alongside latency and cost. Third, transparency and auditability become essential, as users and service providers require visibility into how resources were consumed and how outcomes were produced.

These challenges point toward the need for billing-grade execution tracing, cost-aware orchestration policies, and service contracts that explicitly incorporate Economic QoS (E-QoS) metrics~\cite{zheng2014qos}. In this model, agentic services are evaluated not only by functionality, but by their ability to deliver predictable, auditable, and cost-efficient execution under uncertainty. Recent industry analyses also point to emerging pricing diversity and uncertainty in monetizing agentic workflows, reinforcing the need for principled system-level approaches~\cite{sequoia2024ai,goldman2024ai}.
\section{Future Directions}
\label{sec:future}

The preceding sections describe a progression from protocols, to deployable systems, to service-computing interpretation. The next phase of the IoA is not simply broader adoption; it is the transition from interoperable demonstrations to governable services. Connectivity is increasingly available, but trustworthy autonomy remains unresolved. The key research agenda is therefore to make agent services observable, auditable, and economically controllable enough for sustained production use~\cite{dean2025delegation}. Future directions should therefore be read not as speculative add-ons but as the conditions under which the IoA can become a durable services-computing paradigm.

\subsection{Protocol Maturity}
\label{sec:future-protocols}

Protocol maturity is primarily an operational problem: fewer competing cores, stronger interoperability evidence, and explicit safety guarantees. MCP and A2A are plausible anchors, but long-term value depends on whether they produce dependable behavior across heterogeneous implementations rather than only specification-level agreement~\cite{aaif2025launch,a2a2025spec,anthropic2024mcp,consensus_multiagent}.

\subsubsection{Conformance and Interoperability}
\label{sec:future-conformance}

The immediate requirement is a conformance ecosystem: test suites, interoperability events, and certification processes so that ``MCP-compatible'' or ``A2A-compatible'' implies behavioral interoperability instead of syntax-level similarity~\cite{uddi_mcp_comparison,aaif2025launch}.

\subsubsection{Safety and Verifiable Identity}
\label{sec:future-safety-identity}

Safety and identity must mature together. Formal analysis is needed for liveness, delegation bounds, and policy soundness in long-running workflows, while cross-organizational deployment will likely require stronger, verifiable agent identity and zero-trust interaction patterns beyond today's baseline OAuth and certificate assumptions~\cite{a2a2025spec,consensus_multiagent,agntcy2025}.

\subsection{Systems Maturity}
\label{sec:future-systems}

Systems maturity will come from control-plane quality, not from thicker agent wrappers. Managed runtimes must make traces, policy checkpoints, grounding quality, and cost envelopes first-class operational controls so that long-running multi-agent workloads are governable under production constraints~\cite{dean2025sequoia,facts2025grounding,dean2025delegation}.

\subsubsection{Agent Operating-System Layers}
\label{sec:future-agent-os}

A plausible trajectory is an agent operating-system layer in which compute budgets, memory residency, task scheduling, and inter-agent coordination become native runtime responsibilities. Under this view, lifecycle management and declarative deployment are not convenience features; they are core service infrastructure.

\subsubsection{Grounded Inference and Infrastructure Co-Design}
\label{sec:future-grounded-codesign}

Grounded inference and hardware co-design must also advance together. ``Filter first, reason last'' pipelines reduce failure propagation by improving context quality before expensive reasoning begins, while serving stacks need to adapt from static batch assumptions toward irregular, stateful, tool-using workloads with dynamic branching and context reuse~\cite{lewis2020rag,dean2025sequoia,facts2025grounding}.

\subsection{Services Maturity}
\label{sec:future-services}

Services maturity depends on governance discipline. The operational shift is from querying models to commissioning bounded agents, which requires delegation policies, supervision points, auditability, and cost controls as default service semantics~\cite{dean2025delegation}.

\subsubsection{Enterprise Governance and Commissioning}
\label{sec:future-enterprise-governance}

Enterprise copilots already demonstrate this transition by embedding grounding, identity, retention, audit, and admin governance directly into service environments~\cite{m365copilot_arch2026,m365copilot_data_protection2026,purview_ai2025,m365copilot_connectors2026}. A key remaining challenge is portability: governance controls that work in one tenant need to generalize to cross-organizational delegation and multi-provider workflows. This is where the five requirements articulated by Toma\v{s}ev et al.\---dynamic assessment, adaptive execution, structural transparency, scalable market coordination, and systemic resilience\---become useful as a practical control model for task fit, supervisory depth, and revocation design~\cite{dean2025delegation}.

\subsubsection{Bounded Delegation and Accountability}
\label{sec:future-bounded-delegation}

Cryptographically bounded delegation is a concrete path forward: orchestrators grant narrowly scoped authority to sub-agents instead of broad static credentials, and delegation rights follow task decomposition boundaries~\cite{dean2025delegation,birgisson2014macaroons}.

\begin{figure}[t]
\centering
\includegraphics[width=\linewidth]{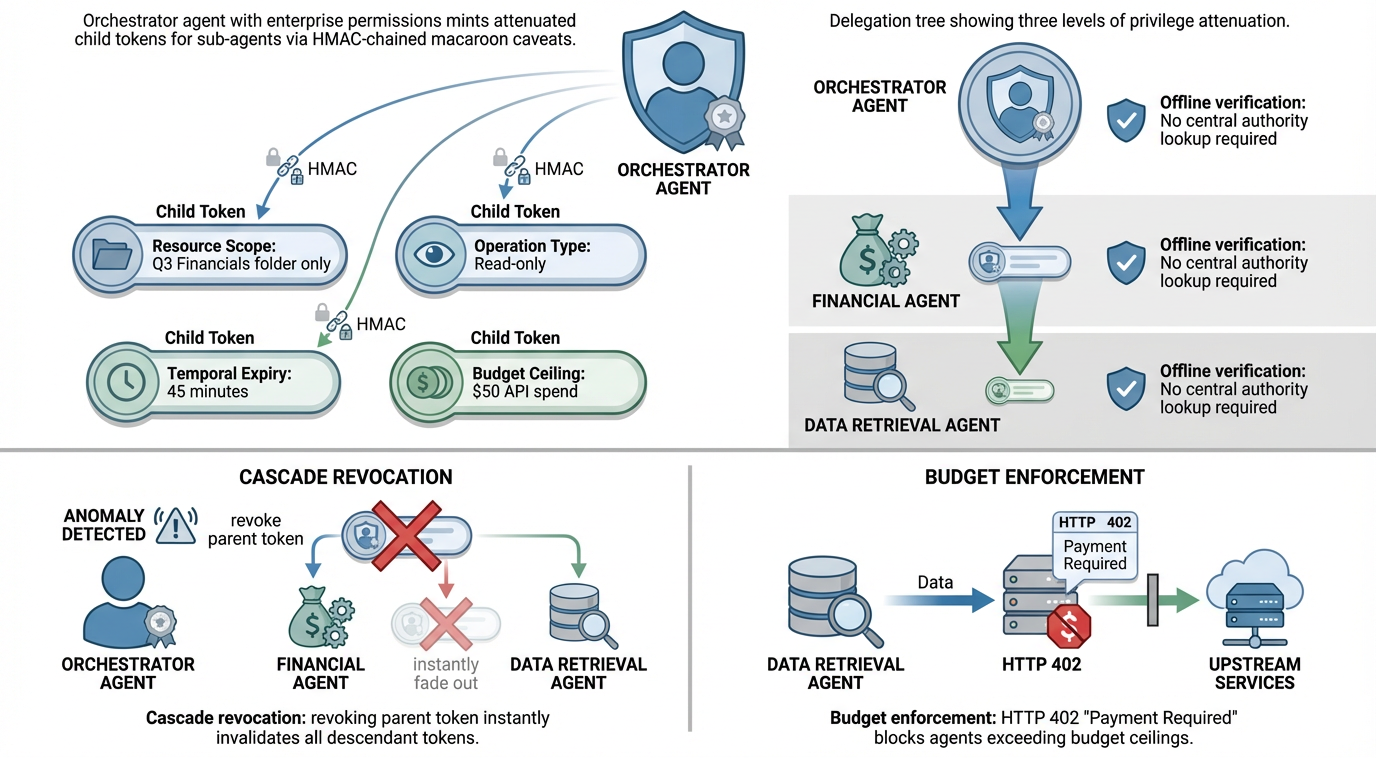}
\caption{Delegation Capability Token architecture for multi-agent trust enforcement, showing privilege attenuation and cascade revocation across delegated workflows.}\vspace{-3mm}
\label{fig:dct-architecture}
\end{figure}

Figure~\ref{fig:dct-architecture} visualizes this idea as a Delegation Capability Token architecture. The figure shows how authority can be attenuated as work moves through a multi-agent task graph, and why revocation must cascade along the same delegation chain if downstream agents inherit limited execution rights from upstream orchestrators.

If agent services are to cross organizational boundaries, authority must be explicit, attenuable, and revocable at task-graph granularity. In parallel, service maturity will require clearer liability chains and auditable accountability in high-stakes domains where no single provider controls full workflow behavior~\cite{a2a2025spec}.

\subsection{Open Research Challenges}
\label{sec:open-challenges}

The open research problems are best framed as service-computing questions rather than as generic LLM limitations. Hallucination, latency, and cost are already familiar concerns in agent deployments, but they become research challenges when they must be translated into quality-of-service (QoS), service-level agreement (SLA), and service-governance terms. Prior work on QoS prediction, recommendation, and probabilistic service selection shows that service quality has long been treated as a measurable, context-dependent contract rather than a single accuracy score~\cite{zheng2011qosrecomm,tang2012locationqos,hwang2015probqos}. A conventional service contract can specify availability, latency percentiles, throughput, error rates, and recovery objectives. An agentic service contract must additionally specify how much uncertainty is tolerable, how validation coverage is measured, how latency budgets are allocated across reasoning and tool calls, and how marginal cost is bounded when the agent can branch, retry, and delegate work.

\subsubsection{Hallucination, Harness Coverage, and Semantic QoS}
\label{sec:challenge-hallucination-qos}

Hallucination is unlikely to be eliminated completely, especially in long-running, multi-agent workflows where retrieved context, intermediate summaries, and delegated outputs can all introduce uncertainty. The services-computing question is therefore not only how to reduce hallucination, but how to deliver acceptable QoS in its presence. Harness engineering provides one practical direction: agents can be surrounded by retrieval checks, schema validators, policy gates, tool-output verifiers, sandbox execution, approval checkpoints, and post hoc audits~\cite{mast2025failures,facts2025grounding,dean2025delegation}. Yet this raises a deeper measurement problem. Unlike code coverage, harness coverage is not simply the fraction of statements or branches exercised; it must describe which semantic claims, tool side effects, delegation paths, policy constraints, and high-risk business states are covered by validation.

This creates a new research agenda for \emph{semantic QoS}: an extension of classical QoS in which contractual quality attributes range over the meaning of outputs (grounding fidelity, factual accuracy, policy conformance) rather than only over observable service attributes such as latency or availability~\cite{facts2025grounding,hwang2015probqos}. Future agentic services need metrics that quantify the coverage of grounding evidence, the sensitivity of outputs to missing or corrupted context, the fraction of delegated actions protected by deterministic checks, and the expected harm of an undetected false claim. These measures must then be mapped to SLA language, extending earlier QoS models from observable service attributes toward evidence-backed semantic behavior~\cite{facts2025grounding,hwang2015probqos}. For example, an enterprise agent may not be able to promise zero hallucinations, but it might promise that claims affecting financial reporting, compliance decisions, or external side effects are backed by auditable evidence and deterministic validation before execution. The unresolved problem is how to make such guarantees portable across heterogeneous agents and protocols rather than bespoke to one application.

\subsubsection{Inference Latency and Agentic SLA Design}
\label{sec:challenge-latency-sla}

Latency is also different for agentic services than for ordinary API calls. A user-facing response may involve multiple LLM invocations, retrieval calls, tool executions, validation passes, and cross-agent delegations. At the infrastructure layer, inference latency depends on GPU availability, prefill and decode behavior, KV-cache residency, cache eviction, GPU--CPU bandwidth, and migration across memory tiers. These details are often hidden from the service consumer, but they directly shape whether an agent can meet an SLA for an interactive task, a batch workflow, or a time-sensitive business process.

The research challenge is to design SLAs that reflect the compositional nature of agent latency. Traditional percentiles such as p95 response time are insufficient by themselves when a task graph may expand dynamically after intermediate reasoning, just as classical service selection had to reason over end-to-end QoS rather than isolated endpoint quality~\cite{hwang2015probqos}. Agentic SLAs need latency budgets that can be decomposed across planning, model inference, tool waiting, validation, and human escalation. They also need runtime policies that can trade quality against time: using a smaller model, reusing cached context, truncating low-value evidence, parallelizing subtasks, skipping optional validation, or escalating to a human when the remaining latency budget is too small. This connects hardware-aware scheduling to services computing: GPU and KV-cache management become QoS controls only when the runtime exposes their effects to the orchestration layer in terms that service contracts can use.

\subsubsection{Cost, Elasticity, and Economic QoS}
\label{sec:challenge-cost-qos}

Cost is the third agentic challenge that must be elevated into service-level semantics. Multi-agent execution makes cost difficult to predict because a single request can trigger variable numbers of model calls, tool invocations, retrieval operations, validation passes, and retries~\cite{google2025scaling,a2a2025spec}. Under a cost-per-use model, this variability is not merely an accounting issue; it changes the architecture of the service. A reliable agentic service must estimate the expected cost of a task before accepting it, monitor cost during execution, and degrade, defer, or reject work when the remaining budget is inconsistent with the requested quality level.

This suggests an economic form of QoS for agentic services. Future systems should expose service contracts that combine functional quality, confidence, latency, and marginal cost rather than optimizing them separately. This direction generalizes QoS-aware service recommendation and selection from choosing among service endpoints to choosing among reasoning, retrieval, tool, and validation paths~\cite{zheng2011qosrecomm,tang2012locationqos,hwang2015probqos}. For example, an agent marketplace may need to compare services by expected dollars per verified outcome, not only by accuracy or uptime. Orchestrators may need cost-aware planning algorithms that decide when extra verification is worth its price, when cheaper models are sufficient, when cached evidence can be reused safely, and when additional delegation creates more cost than value. The unresolved research problem is how to make these choices auditable and contractible: a user should be able to understand why an agent spent a given amount, what quality improvement that spending purchased, and whether the final output satisfied the agreed service objective.

Taken together, hallucination, latency, and cost show why agentic services cannot be evaluated only with single-agent benchmarks~\cite{facts2025grounding}. The field needs benchmarks and contracts that measure end-to-end task graphs under uncertainty: coverage of validation harnesses, latency under realistic GPU and memory constraints, cost under branching delegation, and recovery behavior when the agent violates a policy or exhausts a budget. Cross-protocol interoperability will remain necessary even if the ecosystem partially converges~\cite{ibm2025acp,agentprotocol2024}, but interoperability alone is not enough. The durable research challenge is to turn probabilistic reasoning into service behavior whose quality, delay, and cost can be measured, governed, and promised.

\section{Discussion and Conclusion}
\label{sec:limitations}
\label{sec:conclusion}

This paper presented an integrative architectural perspective on the emerging Internet of Agentics (IoA), synthesizing protocols, systems, and services-computing developments into a unified framework for trustworthy and governable agentic ecosystems. Rather than treating agentic systems as isolated AI applications, the paper argued that the IoA represents a broader evolution of services computing toward interoperable cognitive service infrastructures.

The paper emphasizes architectural synthesis and service-level interpretation over empirical benchmarking, a positioning that aligns with the still-emerging standards landscape while necessarily limiting the scope of empirical claims. Protocol details and production claims often appear first in specifications, public repositories, engineering blogs, and foundation announcements, rather than in peer-reviewed venues. Accordingly, the paper treats such materials as signals of ecosystem direction and architectural intent rather than as conclusive evidence of adoption, performance, or long-term stability. Given the rapid evolution of the field, protocols, governance structures, and enterprise deployments—including MCP, A2A, Agent Cards, and AAIF initiatives—may continue to evolve between manuscript writing and publication~\cite{mcpspec2025,a2a2025spec,aaif2025launch,google2025scaling}. Some recent technical concepts, such as \emph{using Command Line Interface (CLI) to replace MCP}~\footnote{Summary of the MCP vs. CLI debate on X this week.
(Mar 14, 2026), \url{https://x.com/randal_olson/status/2032513098838597714}} and \emph{You Only Live Once (YOLO) for coding agent permission control}~\footnote{[FEATURE] Add YOLO mode toggle support to Claude Code CLI, \url{https://github.com/anthropics/claude-code/issues/33046}}, which are not yet standardized but widely adopted are not discussed here. 

In addition, the service-computing analogies used throughout the paper are analytical aids rather than exact historical equivalences: they clarify recurring problems of discovery, composition, governance, and trust, while the uncertainty introduced by probabilistic reasoning and emergent multi-agent behavior remains distinctive. Legal liability, labor impacts, domain-specific compliance, and controlled benchmarking of protocol implementations require separate treatment as agent services move into high-stakes settings~\cite{dean2025delegation}.

Within these boundaries, the Internet of Agentics should be understood as an effort to transform interoperable agents into trustworthy cognitive services. Its protocols are beginning to standardize how agents access tools and coordinate with peers. Its systems abstractions are beginning to make agents deployable as managed runtime components. Its service implications are best interpreted through the long history of service discovery, orchestration, and governance. The main unresolved issue is not whether agents can act, but whether they can act under constraints strong enough for production environments~\cite{anthropic2024mcp,a2a2025spec,papazoglou2003soa}.

Seen in that light, this perspective paper argues that the central question is shifting from whether distributed autonomy is technically possible to whether it can be made trustworthy enough for production services. The unresolved question is whether distributed autonomy can be transformed into trustworthy autonomy through protocol verification, grounded inference architectures, bounded delegation, stronger runtime governance, and service models that treat agents as accountable actors rather than as stochastic search boxes. That transformation is the condition under which the IoA can become a durable services-computing paradigm rather than a transient integration wave~\cite{dean2025delegation,facts2025grounding,consensus_multiagent}.

\bibliographystyle{IEEEtran}
\bibliography{references}

\end{document}